\documentclass[prep,sglanonrev]{arxiv}

\RequirePackage{tgtermes}
\RequirePackage{newtxtext}
\RequirePackage{newtxmath}
\RequirePackage{bbm}
\RequirePackage{endnotes}
\RequirePackage{rotating}

\OneAndAHalfSpacedXI

\usepackage{tikz}
\usepackage{hyperref}
\usepackage{url}

\usepackage{hyperref}       
\usepackage{url}     
\usepackage{booktabs}       
\usepackage{amsfonts}       
\usepackage{nicefrac}       
\usepackage{microtype}      
\usepackage{xcolor}         
\usepackage{amsmath, amssymb, mathtools}
\usepackage{dsfont}
\usepackage{graphicx}
\usepackage{algorithm, algorithmic}
\usepackage{cleveref}
\usepackage{subcaption}
\usepackage{adjustbox}
\usepackage{threeparttable}
\usepackage{multicol, multirow}

\newcommand*{\eg}{\emph{e.g.}{}}
\newcommand*{\ie}{\emph{i.e.}{}}

\usepackage{natbib}
 \bibpunct[, ]{(}{)}{,}{a}{}{,}%
 \def\bibfont{\small}%

\EquationsNumberedThrough
\TheoremsNumberedThrough
\ECRepeatTheorems

\MANUSCRIPTNO{}

\begin{document}


\RUNAUTHOR{Zhou, Orfanoudaki, and Zhu}

\RUNTITLE{Conformalized Decision Risk Assessment}

\TITLE{Conformalized Decision Risk Assessment}

\ARTICLEAUTHORS{
\AUTHOR{Wenbin Zhou}
\AFF{Heinz College of Information Systems and Public Policy,
Carnegie Mellon University,
\EMAIL{wenbinz2@andrew.cmu.edu}}

\AUTHOR{Agni Orfanoudaki}
\AFF{Sa\"id Business School,
University of Oxford, \EMAIL{agni.orfanoudaki@sbs.ox.ac.uk}}

\AUTHOR{Shixiang Zhu}
\AFF{Heinz College of Information Systems and Public Policy,
Carnegie Mellon University,
\EMAIL{shixianz@andrew.cmu.edu}}
}

\ABSTRACT{
In many operational settings, decision-makers must commit to actions before uncertainty resolves, but existing optimization tools rarely quantify how consistently a chosen decision remains optimal across plausible scenarios. This paper introduces CREDO---Conformalized Risk Estimation for Decision Optimization, a distribution-free framework that quantifies the probability that a prescribed decision remains (near-)optimal across realizations of uncertainty. CREDO reformulates decision risk through the inverse feasible region---the set of outcomes under which a decision is optimal---and estimates its probability using inner approximations constructed from conformal prediction balls generated by a conditional generative model. This approach yields finite-sample, distribution-free lower bounds on the probability of decision optimality. The framework is model-agnostic and broadly applicable across a wide range of optimization problems. Extensive numerical experiments demonstrate that CREDO provides accurate, efficient, and reliable evaluations of decision optimality across various optimization settings.
}




\maketitle

\section{Introduction}
\label{sec:Intro}

Organizations routinely make consequential decisions in the presence of substantial uncertainty \citep{kochenderfer2015decision}. Utilities must plan infrastructure upgrades without knowing future load growth or the intensity of extreme weather events \citep{chen2025global}; hospitals must allocate limited staff and beds before patient needs are realized \citep{kim2015two}; and public agencies must design policies whose impacts depend on uncertain behavioral and socioeconomic responses \citep{zhu2022data}. In such settings, decision-makers are tasked with selecting the best action by solving an optimization problem that will remain effective across a range of possible scenarios. Classical approaches address this challenge by representing uncertainty through forecasts or sampled scenarios and subsequently optimizing with respect to this surrogate representation. Two common instantiations of this paradigm are predict-then-optimize \citep{bertsimas2020predictive, elmachtoub2022smart}, where decisions are optimized against a point forecast, and scenario-based stochastic programming \citep{bertsimas2018robust}, where decisions are chosen to minimize expected cost across empirical samples. These tools have formed the backbone of prescriptive analytics in operations and have been successfully deployed across numerous application domains \citep{bertsimas2021predictions, tian2023smart}.

Despite their widespread use, these approaches provide a limited view of the decision landscape. Many operational problems admit multiple actions that perform nearly equivalently across plausible realizations of uncertainty \citep{topkis1998supermodularity, bertsimas2004price}. For example, several investment plans may harden the electric grid to similar degrees under different storm patterns \citep{lombardi2025near}, or multiple staff schedules may achieve comparable service quality across demand scenarios \citep{decarolis2011using, palmintier2014flexibility}. Yet classical optimization pipelines return only a single recommended action, offering little visibility into whether this choice is robust or whether alternative decisions are nearly as effective \citep{li2024balancing, zhang2025admission}. As a result, decision-makers must often rely on informal heuristics or domain intuition to assess the reliability of the prescribed solution and to understand how frequently it would remain optimal as the environment varies \citep{delarue2025algorithmic}. This gap between the deterministic nature of traditional models and the practical need to evaluate the reliability of decisions highlights the need for new rigorous tools that more directly quantify decision reliability.

Our work addresses this gap by adopting a complementary perspective to the established literature. Rather than collapsing uncertainty into a single deterministic representation, we focus on \emph{quantifying the likelihood that a given decision remains (near-)optimal under different realizations of the environment}.
The central question shifts from ``What does the model recommend?'' to ``How reliably will this prescribed decision perform under varying conditions?''
Under this view, analytical tools serve not as automatic decision engines but as decision-support systems that assess how robust a candidate solution is to underlying data variability. Such information is valuable in operational human-in-the-loop contexts where human judgment, regulatory oversight, or organizational risk tolerance ultimately shape the chosen action \citep{dietvorst2018overcoming,grand2024best}. By quantifying the reliability of a decision, the framework provides interpretable evidence that supports accountability, stakeholder communication, and more informed strategic planning.

\begin{figure}
    \FIGURE{
    \includegraphics[width=1.0\linewidth]{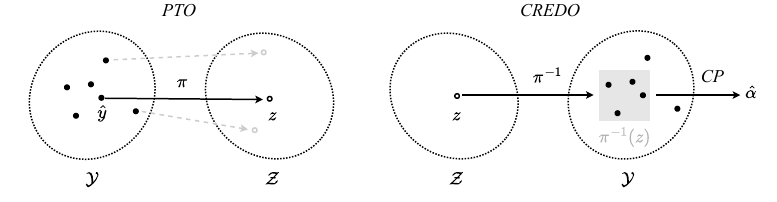}
    }
    {Our proposed framework (CREDO) compared against conventional predict-then-optimize (PTO).}
    {For a given decision $z$ in feasible region $\mathcal{Z}$, CREDO first finds the inverse feasible region $\pi^{-1}(z)$ within scenario space $\mathcal{Y}$ and then estimates the associated decision risk by quantifying the probability that the realized $y$ lies within this region using conformal prediction (CP).
    In contrast, the conventional predict-then-optimize takes an estimated scenario $\hat y$ as input to policy $\pi$ and outputs a single prescribed decision without a reliability assessment. 
    \label{fig:intro-example}}
\end{figure}

To operationalize this perspective, we introduce CREDO---Conformalized Risk Estimation for Decision Optimization, a framework that quantifies, for any candidate decision, a distribution-free lower bound on the probability that it remains (near-)optimal. The proposed approach integrates inverse optimization \citep{chan2025inverse} with distribution-free uncertainty quantification via conformal prediction (CP) \citep{angelopoulos2021gentle} (see Figure~\ref{fig:intro-example} for an illustration). The core challenge is to evaluate the likelihood that an uncertain scenario falls within the decision’s inverse feasible region, namely the set of scenarios under which that decision remains optimal or near-optimal, and directly characterizing this region is generally intractable due to its implicit and complex geometry. 
CREDO circumvents this challenge by constructing inner approximations of the inverse feasible region using multiple CP balls generated from a conditional generative model. Each ball constitutes a valid prediction region for the outcome with guaranteed coverage. By shrinking the radius so that the ball lies entirely within the inverse feasible region, we obtain a certified lower bound on the decision’s optimality probability given by the corresponding coverage level. Averaging these calibrated bounds yields a distribution-free, finite-sample valid estimator of decision risk.
CREDO is compatible with modern generative forecasting models, supports multiple conformal radius constructions, and applies broadly to convex decision problems. We establish theoretical validity for the estimator, develop efficient computational procedures, and demonstrate through extensive numerical experiments that CREDO provides accurate, interpretable, and reliable assessments of decision reliability across various optimization settings. The key contributions of our work are:
\begin{itemize}
    \item \textit{Problem Formulation}: 
    We formulate the decision risk assessment as estimating the probability that a candidate decision remains (near-)optimal under uncertainty. By expressing this probability through inverse feasible regions, we convert an intractable evaluation problem into a tractable probability estimation task.
    \item \textit{CREDO Framework}: 
    We introduce CREDO, a distribution-free framework based on CP that delivers valid and finite-sample lower bounds on decision optimality, and establish marginal conservativeness, asymptotic consistency, and accuracy guarantees.
    \item \textit{Computational Algorithm}:
    We derive a closed-form estimator for linear programs and develop an efficient computational procedure applicable to general convex optimization problems.
    \item \textit{Empirical Validation}:
    Through controlled experiments featuring various optimization paradigms such as linear, quadratic, and second-order conic programs, as well as a real-world power grid planning application, we verify CREDO's theoretical guarantees and demonstrate its practical advantages. Our method achieves 100\% empirical validity under conservative radius choices, superior true positive rates (up to 78.75\% improvement over baselines), consistently selects decisions with higher empirical confidence rankings than existing prescriptive methods, and is superior compared to various ablation variants.
\end{itemize}

The remainder of this paper is organized as follows. Section~\ref{sec:literature} reviews the related literature and positions our contributions. Section~\ref{sec:setup} formalizes the decision risk assessment problem and provides motivating examples. In Section~\ref{sec:method}, we present the CREDO methodology, combining inverse optimization with generative CP. Section~\ref{sec:theory} establishes theoretical guarantees including validity, consistency, and true positive rate analysis. Section~\ref{sec:computational} details the proposed computational implementation of CREDO, while Section~\ref{sec:experiments} demonstrates its effectiveness through three numerical experiments. Section~\ref{sec:conclusion} concludes with discussions of implications and future work. Technical proofs and additional experiment details are included in the Appendix.

\section{Related Work}
\label{sec:literature}

Our work draws from and contributes to four distinct research streams: ($i$) decision-making under uncertainty in operations research, ($ii$) distribution-free uncertainty quantification through CP, ($iii$) inverse optimization for understanding decision structures, and ($iv$) integration of CP and human-in-the-loop in robust decision making. In what follows, we position the CREDO approach relative to these lines of research.

Classical approaches to optimization under uncertainty can be categorized by their treatment of unknown parameters. Stochastic optimization seeks decisions that minimize the expected value of the objective function across probabilistic scenarios, typically implemented through sample average approximation over historical data \citep{shapiro2021lectures, kleywegt2002sample,
lan2025bias}. The predict-then-optimize (PTO) paradigm offers an alternative by first estimating unknown parameters via predictive models, then solving the resulting deterministic optimization problem \citep{bertsimas2020predictive, elmachtoub2022smart}. Acknowledging that probability distributions themselves may be misspecified, robust optimization (RO) is employed to hedge against worst-case realizations within specified uncertainty sets \citep{bertsimas2006robust,ben2002robust}. Distributionally robust optimization (DRO) generalizes this concept further, considering worst-case probability distributions rather than point realizations, thereby balancing conservativeness with statistical plausibility \citep{delage2010distributionally, rahimian2022frameworks}. Recognizing that prediction errors can compound when models are trained separately from their downstream use, decision-focused learning (DFL) trains predictive models end-to-end by differentiating through the optimization layer, directly minimizing downstream decision costs rather than prediction errors \citep{amos2017optnet, mandi2024decision}. While these prescriptive paradigms effectively recommend decisions, our work adopts a parallel goal: rather than \textit{prescribing} a decision, we focus on \textit{auditing} decisions by providing a quantitative, data-driven assessment of the risk associated with taking a given decision.
This perspective complements existing decision-prescription methods and even human judgment by adding an additional layer of transparency and accountability, thereby supporting more reliable and defensible decision-making.

Methodologically, our work builds upon the conformal prediction (CP) literature, a distribution-free approach for uncertainty quantification through calibrated prediction sets \citep{shafer2008tutorial, vovk2005algorithmic}.
Multiple strands of recent research are relevant to our work. First, generative conformal methods draw multiple samples from generative models to construct both the calibration sets and the resulting CP regions, thus significantly improving efficiency (\ie, tightness of the prediction sets) especially when the data distribution is highly complex or dispersed \citep{zheng2024optimizing, zhou2024hierarchical}. Second, inverse CP estimates miscoverage rates for fixed prediction sets by identifying the smallest miscoverage level at which conformal regions are contained within the target set \citep{prinster2023jaws, singh2024distribution}. Advances in $e$-value CP strengthen this approach by enabling post-hoc selection of the miscoverage rate while preserving finite-sample validity at the theoretical level \citep{vovk2025conformal, gauthier2025values}. Our algorithm synthesizes both advances by leveraging generative models to enhance decision-risk estimation accuracy and subsequently employing generalized inverse CP to assess the decision risk. However, we emphasize that our problem setting is fundamentally different and cannot be addressed by direct application of these methods. Substantive reformulation is required to adapt their underlying ideas to the decision-auditing context.

Our work also connects to inverse optimization (IO), which seeks to infer the unknown parameters of an optimization problem from observed solutions \citep{ahuja2001inverse, chan2025inverse, aswani2018inverse, zattoni2025learning, besbes2025contextual}. Recent work on Conformal-IO \citep{lin2024conformal, chan2024conformal} combines IO with CP to prescribe robust decisions by first estimating decision-makers' preference parameters from observed choices and then applying robust forward optimization using CP to construct an uncertainty set over these parameters. While we similarly integrate IO and CP, our learning setting fundamentally differs. We assume access to samples of the unknown parameters rather than the actual decisions. This leads us to employ inverse CP to assess decision reliability \citep{singh2024distribution} rather than forward CP to prescribe decisions. Our focus on evaluation rather than prescription distinguishes our approach from Conformal-IO's goal of generating policy recommendations. Meanwhile, though efficient IO algorithms exist for structured problems like linear, integer, and convex programs \citep{tavasliouglu2018structure, schaefer2009inverse, iyengar2005inverse}, the lack of a unified framework for general IO leads to the adoption of structured formulations for tractability \citep{chan2024conformal}. This motivates our focus on linear models as part of our theoretical analysis, allowing us to derive an efficient closed-form solution for the proposed implementation algorithm.

Finally, our research complements a growing line of work on the integration of CP to robust decision-making \citep{kiyani2025decision, patel2024conformal, cortes2024decision, yeh2024end, hullman2025conformal, cresswellconformal2, zhao2025conformal}.
Unlike prior work that prescribes decisions, we leverage CP techniques to audit and assess the risk of candidate decisions, supporting decision-making through evaluation rather than recommendation. Additionally, our work contributes to the growing literature on human-algorithm collaboration through the use of generative model recommendations \citep{grand2024best, ibrahim2021eliciting, orfanoudaki2022algorithm}. By producing a diverse set of plausible decisions, generative algorithms can act as advisory tools, thus promoting exploration and helping users develop more effective decision strategies \citep{ajay2022conditional, li2024balancing}.

Aligned with this perspective, our approach encourages decision-makers to evaluate alternatives beyond standard prescriptive recommendations, revealing risks and improvements that the nominal optimal decision alone may obscure. This emphasis on expanding the space of considered decisions is a central conceptual contribution of our work.

\section{Problem Setup}
\label{sec:setup}

We consider a general parametric decision-making model in which the decision is determined as the (near-)optimal solution to a constrained optimization problem under an observed scenario $y \in \mathcal{Y}$ (\eg, demand, price, or system load). 
Formally, we define the \emph{$\epsilon$-optimal decision rule} as
\begin{equation}
    \label{eq:epsilon-solution-set}
    \pi_\epsilon(y)
    \coloneqq
    \arg\min\nolimits_{\epsilon,~ z \in \mathcal{Z}}
    f(z; y),
\end{equation}
where $f$ denotes the objective function (\eg, cost, delay, or loss), and $\mathcal{Z}$ denotes the feasible region. 
We assume that $\mathcal{Z}$ is nonempty and compact, and that $f(\cdot; y)$ is continuous and convex in $z$ for each $y$, ensuring the set $\pi_\epsilon(y)$ is non-empty for every $y \in \mathcal{Y}$.
The operator $\arg\min\nolimits_\epsilon$ generalizes the standard $\arg\min$ to allow a tolerance level $\epsilon \ge 0$, defined as:
\[
    \arg\min\nolimits_\epsilon f(z)
    ~\coloneqq~
    \left\{
        z ~\middle|~
        f(z) \le \inf_{z'} f(z') + \epsilon
    \right\}.
\]
Note that setting $\epsilon = 0$ recovers the \emph{exact optimal decision rule}. Throughout the manuscript, we suppress the subscript $\epsilon$ when no ambiguity arises, writing $\pi(y)$ for $\pi_\epsilon(y)$.

In many operational contexts, the realized scenario $y$ is unobserved at the time of decision-making. 
Let $z \in \mathcal{Z}$ denote a prescribed (implemented) decision---potentially produced by a human operator, heuristic policy, or black-box model---made without access to the true realization of $Y \sim \mathcal{P}_Y$. 
This setup captures a common class of practical decision environments where actions are based on implicit or subjective beliefs about the underlying uncertainty rather than a fully specified probabilistic model.

Our objective is to assess how well a prescribed decision $z$ aligns with the (unknown) optimal response under the true realization of $Y$. 
Formally, this quantity is expressed as the probability that $z$ belongs to the (near-)optimal decision set $\pi(Y)$, \ie, $\mathbb{P}\{ z \in \pi(Y) \}$, where the probability is taken with respect to $Y \sim \mathcal{P}_Y$. 
This captures how frequently the prescribed decision coincides with an (approximately) optimal choice across possible realizations.
Since the distribution $\mathcal{P}_Y$ is typically unknown and only limited samples $\mathcal{D} = \{Y_i\}_{i=1}^n$ are available, this probability cannot be estimated precisely. We therefore focus on its data-driven lower bound $1- \alpha(z)$ defined as
\begin{equation}
    \label{eq:risk-certificate}
    \mathbb{P}\bigl\{ z \in \pi(Y) \bigr\} \ge 1 - \alpha(z).
\end{equation}
The quantity $1 - \alpha(z)$ serves as a conservative optimality certificate, providing a guaranteed confidence level that the prescribed decision $z$ is (near-)optimal under the true, unknown distribution $\mathcal{P}_Y$. 

In the general setting, the quantities in \eqref{eq:risk-certificate} may depend on additional contextual covariates $X \in \mathcal{X}$, such as spatial, temporal, or environmental factors that influence the realization of $Y$. 
Accordingly, both the decision set $\pi(Y)$ and the confidence function $\alpha(z)$ can be defined conditionally on $X$, reflecting the dependence of the outcome distribution $\mathcal{P}_{Y|X}$ on its context. 
Also, the available data take the form $\mathcal{D} = \{(Y_i, X_i)\}_{i=1}^n$ in this case, capturing paired observations of outcomes and their associated covariates.
For clarity of exposition, in the remainder of the paper, we omit the explicit dependence on $X$ and focus on the marginal representation.

\subsection{Motivating Examples}

We provide two illustrative examples to demonstrate how the proposed problem setup applies to different decision-making contexts, ranging from simple binary decisions to complex resource allocation problems.

\begin{example}[Stylized Umbrella-Carrying Decision]
    A human decision-maker decides whether to carry an umbrella without checking the weather. 
    Let $z \in \{0,1\}$ denote the decision ($z=1$: carry; $z=0$: not carry) and $y \in \{0,1\}$ denote the realized weather ($y=1$: rain; $y=0$: no rain). 
    The feasible set is $\mathcal{Z}=\{0,1\}$, and the objective function is a hinge-type misclassification cost:
    \[
    f(z;y) = 
    \begin{cases}
    1, & \text{if } z \neq y \text{ (mismatch between decision and weather)},\\[2pt]
    0, & \text{otherwise,}
    \end{cases}
    ~\text{or simply}~
    f(z;y) = \mathbbm{1}\{z \neq y\}.
    \]
\end{example}

In this example, the decision-maker acts based on an implicit perception of local weather patterns $Y$. 
The chosen decision $z$ thus reflects the decision-maker’s subjective belief about the likelihood of rain---carrying an umbrella if rain is perceived as more probable, and abstaining otherwise. 
Our goal is to assess whether this private knowledge aligns with the true local weather distribution by evaluating, through empirical data, how frequently the prescribed decision coincides with the optimal decision under the observed realizations of $Y$.

\begin{example}[Clinical Triage]
    A clinician must allocate limited treatment resources under time pressure without reviewing complete patient records. Let there be $d$ patients, and define the decision $\mathbf{z} = (z_1,\dots,z_d)\in \{0,1\}^d$, where $z_j = 1$ if patient $j$ receives treatment, subject to a capacity constraint $\sum_{j=1}^d w_j z_j \le B$ (\eg, limited ICU beds, clinician time, or medication supply), with $w_j > 0$ denoting resource use of patient $j$ and $B > 0$ the available budget. Let $\mathbf{x} = (x_1, \dots, x_d)$ represent observable clinical covariates, and $\mathbf{y} = (y_1, \dots, y_d)$ with $y_j \ge 0$ the realized benefit of treating patient $j$ given $x_j$. The feasible region and objective function are defined as
    \[
    \mathcal{Z} = \Bigl\{ \mathbf{z} \in \{0,1\}^d : \sum_{j=1}^d w_j z_j \le B \Bigr\}, 
    \quad
    f(\mathbf{z}; \mathbf{y}) = -\sum_{j=1}^d y_j z_j.
    \]
\end{example}

In this more realistic setting, the clinician makes treatment decisions based on incomplete information or heuristic prioritization, such as triage scores or observable symptoms \citep{parenti2014}.
The prescribed decision vector $\mathbf{z}$ encodes the clinician’s implicit belief about which patients are most likely to benefit given the resource constraint. 
We aim to evaluate whether such implicit prioritization aligns with the optimal clinical decisions suggested by data, by comparing the prescribed allocation with the optimal allocation derived from empirical patient records.

\section{The CREDO Framework}
\label{sec:method}

We now present our method, which adapts the distribution-free uncertainty quantification framework of CP \citep{vovk2005algorithmic} to estimate decision risk through calibrated inverse feasible regions.

\subsection{Preliminary: Conformal Prediction}
\label{sec:cp}

CP is a model-agnostic framework for uncertainty quantification that delivers finite-sample valid prediction regions under the exchangeability assumption:
\begin{assumption}[Exchangeability]
    \label{ass}
    The calibration data $\{(X_i,Y_i)\}_{i=1}^n$ and the test point $(X,Y)$ are exchangeable. 
    That is, the joint distribution of 
    $
    (X_1,Y_1),\dots,(X_n,Y_n),(X,Y)
    $
    is invariant under any permutation of these $(n+1)$ pairs.
\end{assumption}
Let the calibration dataset be $\mathcal{D} = \{(X_i, Y_i)\}_{i=1}^n$, drawn from an arbitrary joint distribution over $\mathcal{X} \times \mathcal{Y}$. 
For a user-specified miscoverage rate $\alpha$, the objective is to construct a prediction set $C(\cdot; \alpha)$ such that
\begin{equation}
    \label{eq:obj-naive}
    \mathbb{P}\left\{ Y \in \mathcal{C}(X; \alpha) \right\} \ge 1 - \alpha,
\end{equation}
where the probability is taken jointly over the randomness in the test point $(X, Y)$ and the calibration data $\mathcal{D}$. This requirement is referred to as \textit{marginal validity}.

Split CP \citep{papadopoulos2002inductive} provides a practical and computationally efficient method for achieving the guarantee in \eqref{eq:obj-naive}. The procedure begins by training a prediction model $g : \mathcal{X} \to \mathcal{Y}$ using data independent of the calibration set. We then specify a nonconformity score function
$l : \mathcal{X} \times \mathcal{Y} \to \mathbb{R}$,
which quantifies how atypical a candidate label is relative to the model’s prediction; for instance, one may take $l(x,y) = \|y - g(x)\|_2$.
Given the calibration dataset, we compute the corresponding nonconformity scores 
\[
    L_i = l(X_i, Y_i), \quad \forall i=1, \dots, n.
\]
These scores determine how large a deviation from the model is acceptable in order to guarantee the desired coverage. For any $\alpha \in [1/(n+1),1)$, we define the adjusted empirical quantile as
\[
\hat{Q}(\alpha)
= \inf\left\{
    l \in \mathbb{R}
    : \hat{F}_n(l) \ge \frac{\lceil (n+1)(1-\alpha) \rceil}{n}
  \right\},
\quad\text{where}\quad
\hat{F}_n(l) = \frac{1}{n}\sum_{i=1}^n \mathbbm{1}\{L_i \le l\}.
\]
This quantile defines the prediction region:
\[
\mathcal{C}(x;\alpha)
= \left\{ y \in \mathcal{Y} : l(x,y) \le \hat{Q}(\alpha) \right\}.
\]
See \citet{angelopoulos2021gentle} for a comprehensive survey of CP.

\begin{figure}[!t]
    \FIGURE
    {
    \includegraphics[width=1.\linewidth]{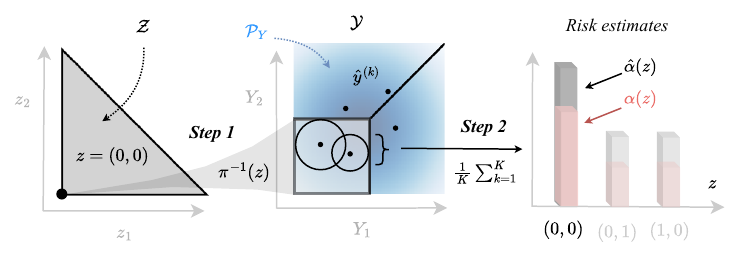}
    \vspace{-1ex}
    }
    {Example of the CREDO procedure applied to a linear programming problem.}
    {There are two main steps:
    ($i$) Map the given candidate decision $z$ (\eg, $(0, 0)$) to its inverse feasible region $\pi^{-1}(z)$;
    ($ii$) Assess the risk by constructing inner approximation of $\pi^{-1}(z)$ using generative conformal prediction, and averaging their miscoverage levels.
    \label{fig:pipeline}
    }
\end{figure}

\subsection{Conformalized Decision Risk Assessment}

We now develop our framework for estimating the decision risk $\alpha(z)$ defined in Section~\ref{sec:setup}. We accomplish this by reformulating the optimality probability in \eqref{eq:risk-certificate} as an \emph{inverse} CP problem, where we seek to quantify how often uncertain parameters fall within regions that render a decision optimal. The key idea is to project the decision's optimality condition onto an \emph{inverse feasible region} \citep{chan2025inverse, tavasliouglu2018structure} in the space of $\mathcal{Y}$, and to construct inner geometric approximations of this region using multiple generated conformal balls \citep{shafer2008tutorial, wang2023probabilistic}.
The resulting average miscoverage rate across balls then serves as a data-driven lower-bound of decision risk. We detail this procedure in two steps, with an example illustrated in \Cref{fig:pipeline}.

\subsubsection{Reformulation with Inverse Feasible Region}

Our first step reformulates \eqref{eq:risk-certificate} as an inverse optimization problem. For any given realization $y$, a decision $z$ is (near-)optimal if and only if it attains the (near-)minimal objective value among all feasible decisions.
Accordingly, we define \textit{inverse feasible region} as the set of scenarios $y$ under which $z$ remains the (near-)optimal decision. Formally,
\begin{equation}
    \label{eq:inv-feasible-set}
    \pi^{-1}_\epsilon(z) \coloneqq \left\{ y \in \mathcal{Y} ~\middle|~ 
    f(z; y) \leq \min_{z' \in \mathcal{Z}} f(z'; y) + \epsilon \right\}.
\end{equation}
This definition can be viewed as a generalization of its classical notion defined by \cite{chan2025inverse}, providing a relaxed definition that accommodates near-optimal decisions.
For notation simplicity, we use $\pi^{-1}$ to denote $\pi_{\epsilon}^{-1}$ when clear from context.
The reformulation is formally stated as follows.

\begin{lemma}[Reformulation]
    \label{lemma:risk-reformulation}
    The probability in \eqref{eq:risk-certificate} can be equivalently expressed as:
    \begin{equation}
        \mathbb{P}\left\{ z \in \pi(Y) \right\} \equiv \mathbb{P}\left\{ Y \in \pi^{-1}(z) \right\}.
    \end{equation}
\end{lemma}
Its proof follows immediately by the definition of $\pi(Y)$ and $\pi^{-1}(z)$. 
This equivalence has important computational implications: the original formulation in \eqref{eq:risk-certificate} requires computing the probability that a fixed decision $z$ lies in the random (near-)optimal set $\pi(Y)$. This task remains computationally intractable because it entails solving the underlying optimization problem for every possible realization of $Y$.
By contrast, the reformulation in \Cref{lemma:risk-reformulation} reverses this viewpoint by evaluating the probability that the \emph{random} variable $Y$ lies within the \emph{deterministic} region $\pi^{-1}(z)$. This shift yields a tractable probability estimation problem.

\subsubsection{Risk Estimation via Generative Conformal Prediction}
The second step estimates the probability $\mathbb{P}\{ Y \in \pi^{-1}(z) \}$.
Direct estimation is generally intractable, as it requires computing the probability mass over $\pi^{-1}(z)$---a region with complex and implicitly defined geometry. Such estimation typically demands either strong parametric assumptions on the distribution of $Y$ or efficient sampling access to it, neither of which is often feasible in practice. A common workaround is to train a conditional generative model that samples from $\mathcal{P}_{Y|X}$ and then approximate the desired probability by the fraction of these samples falling within $\pi^{-1}(z)$. However, this approach lacks any finite-sample guarantee for the validity requirement in \eqref{eq:risk-certificate}.

To address this, we develop a generative conformal approach that constructs statistically valid inner approximations of $\pi^{-1}(z)$ using calibration data.
Given an input $X$, we generate a collection of CP balls $\mathcal{C}(X;\alpha)$ such that ($i$) each region lies entirely within $\pi^{-1}(z)$ and ($ii$) its coverage probability is lower-bounded with statistical guarantees obtained from calibration. This yields the following bound on decision risk:
\begin{equation*}
    \mathbb{P}\left\{ Y \in \pi^{-1}(z) \right\} 
    \stackrel{(i)}{\geq} \mathbb{P}\left\{ Y \in \mathcal{C}(X;\alpha) \right\}
    \stackrel{(ii)}{\geq} 1 - \alpha.
\end{equation*}
The right-hand guarantee follows from the finite-sample validity of CP. To satisfy the left-hand inequality, we identify the smallest $\alpha$ for which the corresponding ball $\mathcal{C}(X;\alpha)$ remains entirely within $\pi^{-1}(z)$. The resulting average value of $\alpha$ across generated balls provides a data-driven estimator of the decision risk, denoted by $\hat{\alpha}(z)$.

\subsubsection{Proposed Algorithm}

We begin with training a (conditional) generative model $g: \mathcal{X} \to \mathcal{Y}$ on a training dataset to approximate the conditional distribution $\mathcal{P}_{Y|X}$.
Then, for a test input $x$, we draw a prediction $\hat y \sim g(x)$ and construct a prediction set as an $\ell_2$ ball centered at $\hat y$:
\begin{equation}
    \label{eq:conformal-set}
    \mathcal{C}(x;\alpha) 
    =
    \left\{ y \in \mathcal{Y} \mid 
    \| y - \hat{y} \|_2 < R(\alpha)
    \right\},
\end{equation}
where the radius $R(\alpha)$ is obtained by calibrating the nonconformity scores $\{L_i\}_{i=1}^{n}$ on $\mathcal{D}$. Of note, $R(\cdot)$ must be a decreasing function, meaning that a smaller $\alpha$ corresponds to a larger $R(\alpha)$ and higher empirical coverage. We consider three specifications of the radius function $R(\alpha): [0,1] \to \mathbb{R}_+ \cup \{\infty\}$, each by default satisfies the boundary conditions $R(\alpha) = \infty$ for $\alpha \in [0, 1/(n+1))$ and $R(1) = 0$. For $\alpha \in [1/(n+1), 1)$, define:
\begin{align}
    R_p(\alpha) &= \hat{Q}(\alpha); && \text{($p$-value radius)} \label{eq:p-radius}\\
    R_e(\alpha) &= \frac{\sum_{i=1}^n L_i}{\alpha(n+1)-1}; && \text{($e$-value radius)} \label{eq:e-radius}\\
    R_\infty(\alpha) &= \infty. && \text{(Monte Carlo radius)} \label{eq:inf-radius}
\end{align}
We note that $R_p$ corresponds to the classical conformal radius based on the empirical quantiles of calibration residuals \citep{vovk2005algorithmic, singh2024distribution};  
$R_e$ is an $e$-value–based variant \citep{grunwald2024safe}, offering stronger post-hoc validity guarantees \citep{vovk2025conformal, balinsky2024enhancing, gauthier2025values};  
and $R_\infty$ is used in a Monte Carlo-based estimation.
We will discuss their respective properties in \Cref{sec:theory} and \Cref{sec:experiments}, showing that $R_e$ provides the strongest robustness, $R_\infty$ the highest accuracy, and $R_p$ a balanced trade-off between the two. 
\begin{figure}[b]
    \FIGURE{\includegraphics[width=1.0\linewidth]{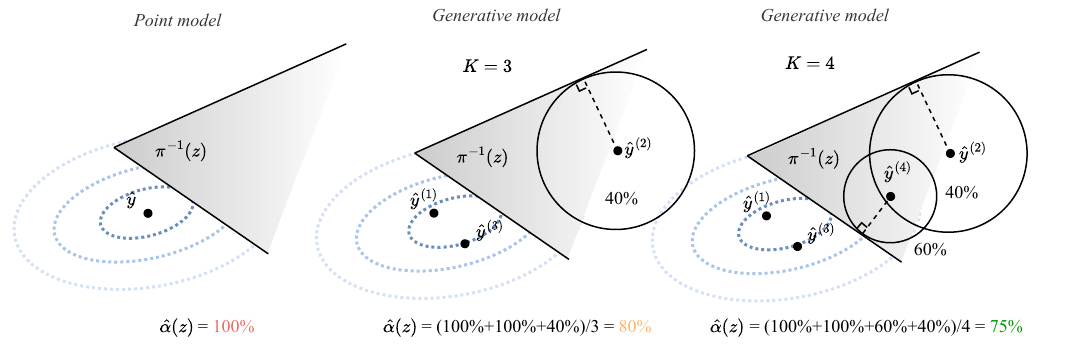}}
    {
    Comparison of risk estimation using a point model and generative models ($K = 3$ and $K = 4$) in CREDO.
    }
    {
    The gray shaded area represents the inverse feasible region $\pi^{-1}(z)$.
    The blue dotted ellipsoids represent the conditional distribution of $Y \mid X$, and the black dots indicate model predictions $\hat y^{(k)}$.
    The balls indicate calibrated prediction sets $\mathcal{C}^{(k)}(x;\hat{\alpha}(z))$.
    \label{fig:caillustration}
    }
\end{figure}

\begin{algorithm}[!t]
\caption{Conformalized Decision Risk Assessment (CREDO)}
\label{alg}
\begin{algorithmic}[1]
    \REQUIRE Fitted generative model $g$; 
    Calibration dataset $\left\{ (x_i, y_i) \right\}_{i=1}^{n}$; Sample size $K$; Decision $z$; Test covariate $x$;
    \STATE Initialize nonconformity score set $\mathcal{L} \leftarrow \emptyset$;
    \FOR{$i \in \left\{ 1, \ldots, n \right\}$}
        \STATE 
        $\hat y_i \sim g(x_i)$
        ;~~$L_i \leftarrow \Vert y_i - \hat y_i \Vert_2$
        ;~~$\mathcal{L} \leftarrow \mathcal{L} \cup \left\{ L_i \right\}$;
    \ENDFOR
    \FOR{$k = 1, \ldots, K$}
        \STATE 
        $\hat y^{(k)} \sim g (x)$ generate prediction using $x$; 
        \STATE $\mathcal{C}^{(k)}(x;\alpha)\leftarrow$ construct conformal set given $\mathcal{L}$ and $\hat y^{(k)}$ via \eqref{eq:conformal-set};
        \STATE $\tilde \alpha^{(k)}(z) \leftarrow$ solve for the $k$-th decision risk via \eqref{eq:alpha-raw};
        \label{step:risk}
    \ENDFOR
    \STATE $\hat \alpha(z) \leftarrow {1} / {K} \cdot \sum_{k=1}^K \tilde \alpha^{(k)}(z)$;
    \RETURN Risk estimator $\hat \alpha(z)$.
\end{algorithmic}
\end{algorithm}

For all variants, the coverage level of each generated prediction set is determined by solving
\begin{equation}
    \label{eq:alpha-raw}
    \tilde{\alpha}(z) = 
        \min_{\alpha \in [0,1]} \alpha
        \quad \text{s.t.} \quad 
        \mathcal{C}(x;\alpha) \subseteq \pi^{-1}(z).
\end{equation} 
The entire procedure above is repeated across $K$ sets to obtain the collection of estimates $\{ \tilde \alpha^{(k)}(z)\}_{k=1}^K$, which are then averaged to yield the final risk estimator:
\begin{equation}
    \label{eq:final-est}
    \hat \alpha(z) = \left( \tilde \alpha^{(1)}(z) + \ldots + \tilde \alpha^{(K)}(z) \right) / K. 
\end{equation}

We highlight the necessity of using generative models in estimating \eqref{eq:final-est} by contrasting them with traditional point prediction models commonly used in standard CP.  
As illustrated in Figure~\ref{fig:caillustration}, point prediction models may produce outputs that lie near the boundary of, or even outside, the inverse feasible region $\pi^{-1}(z)$, leading to overly conservative risk estimates (\ie, extremely small or even zero values when predictions fall outside the region).  
In contrast, generative models enable multiple stochastic draws; increasing $K$ improves the likelihood that at least one sample lies within $\pi^{-1}(z)$, yielding more accurate and less conservative risk estimates.  
This insight will be formalized through the notion of the true positive rate in our theoretical analysis (\Cref{sec:theory}).
The complete algorithmic procedure is summarized in \Cref{alg}.

\section{Theoretical Analysis}
\label{sec:theory}

In this section, we now establish the statistical properties of our risk estimator $\hat{\alpha}(z)$, providing three main theoretical results. First, we show that our estimator is \emph{conservative} (\Cref{thm:val} and \Cref{cor:p}), ensuring it provides a valid upper bound on the true decision risk.
Second, we prove that the estimator is asymptotically consistent (\Cref{prop:consistency}), admitting a tight characterization for the decision risk in large samples.
Finally, we highlight the pivotal role of the generative model in our framework by analyzing the estimator’s true positive rate (\Cref{prop:tpr}), which directly influences the quality of downstream decisions.
All proofs are deferred to Appendix~\ref{app:proof}.

\subsection{Validity}
A fundamental requirement for any risk assessment tool is that it provides reliable guarantees. We therefore first establish that CREDO's risk estimates are conservative for both the $e$- and $p$-value radii, ensuring its trustworthiness to the practitioners. 

\begin{theorem}[Conservativeness under $e$-value]
\label{thm:val}
Under \Cref{ass} and $R(\alpha)=R_e(\alpha)$, the estimator $\hat{\alpha}(z)$ defined in \eqref{eq:final-est} satisfies
\[
\mathbb{P}\{ z \in \pi(Y)\} \ge 1 - \mathbb{E}[\hat{\alpha}(z)].
\]
\end{theorem}
Thus, $\hat{\alpha}(z)$ provides an expectation-wise upper bound on the true decision risk, without requiring correctness of the generative model. This highlights the robustness of the $e$-value construction.
On the other hand, the $p$-value radius has the following guarantee.

\begin{corollary}[Conservativeness under $p$-value]
\label{cor:p}
Under \Cref{ass} and $R(\alpha)=R_p(\alpha)$, the estimator $\hat{\alpha}(z)$ defined in \eqref{eq:final-est} satisfies
\[
\mathbb{P}\{ z \in \pi(Y)\}
\ge
1 - \mathbb{E}[\hat{\alpha}(z)]
 - \frac{1}{n+1} \sum_{i=1}^n d_{\mathrm{TV}}^{(i)}(\hat{\alpha}(z)),
\]
where $d_{\mathrm{TV}}^{(i)}(\hat{\alpha}(z))$ is the conditional total variation (TV) distance between the data vector and the same vector with entries $i$ and $(n+1)$ swapped.
\end{corollary}
Here, the additional TV term reflects the deviation from full exchangeability caused by post-hoc selection of $\hat{\alpha}(z)$ \citep{barber2023conformal,gauthier2025backward}, which disrupts the conservativeness of $p$-value radius's risk estimate but was circumvented by $e$-value radius.
However, we show in our experiments (\Cref{sec:exp1}) that conservativeness for $p$-value radius does hold in practice, and its accuracy is also superior compared to $e$-value radius.
This implies that the additive term in \Cref{cor:p} is usually small in practice, making $p$-value radius a more effective choice when a theoretical guarantee is not the top priority.

\subsection{Consistency}

We next study whether $\hat{\alpha}(z)$ converges to the true risk under mild assumptions on the generative model, measured through total variation distance $d_{\rm TV}(\cdot,\cdot)$. 
The result is summarized as follows.

\begin{proposition}[Asymptotic conditional consistency]
\label{prop:consistency} Let $\hat{\mathcal P}_{Y|X}$ be the conditional distribution learned by the generative model $g$. 
    If for some $\delta \ge 0$,
    \[
    d_{\rm TV}\left(\hat{\mathcal P}_{Y|X=x}, \mathcal P_{Y|X=x}\right) \le \delta
    \quad\text{for all } x \in \mathcal X,
    \]
    then with $R(\alpha)=R_\infty(\alpha)$, the estimator \eqref{eq:final-est} satisfies
    \[
    \left|\hat \alpha(z) - \mathbb{P}\{z \notin \pi(Y)\mid X=x\}\right|
    \le
    O_p(K^{-1/2}) + \delta.
    \]
\end{proposition}
\Cref{prop:consistency} guarantees marginal consistency as $K\to\infty$, provided that the generative model approximates the conditional distribution within TV distance $\delta$.  
Thus, with a well-trained generative model and sufficiently many generated samples, $\hat{\alpha}(z)$ converges at a parametric rate to the true decision risk, complementing the finite-sample conservativeness of \Cref{thm:val}.

\subsection{True Positive Rate}

Finally, we analyze how often the estimator correctly retains feasible decisions.  
Define the true positive rate as
\begin{equation}
    \label{eq:tpr}
    \mathrm{TPR}(K)
    \coloneqq
    \frac{
    \mathbb{E}\left[\#\{ z\in\mathcal Z : \alpha(z)<1 ~\text{and}~\hat \alpha(z)<1\}\right]
    }{
    \#\{z\in\mathcal Z : \alpha(z)<1\}},
\end{equation}
where $\alpha(z)=\mathbb{P}\{z\notin \pi(Y)\}$ is the true risk.
TPR measures the fraction of genuinely feasible decisions that are not mistakenly discarded by the estimator. From a practical perspective, TPR quantifies the ability of CREDO to correctly identify decisions that have non-trivial optimality probability while maintaining conservative guarantees. A low TPR corresponds to more ``false-positive'' exclusions, which may eliminate profitable or optimal decisions; thus, a higher TPR indicates better decision quality.
We have the following result.

\begin{proposition}[True positive rate]
\label{prop:tpr}
The TPR$(K)$ increases monotonically with $K$.
\end{proposition}
\Cref{prop:tpr} highlights the importance of using generative models and of sampling diversity: small $K$ (\eg, $K=1$) or deterministic predictors increase the likelihood of false exclusions, lowering TPR.  
Increasing $K$ mitigates this issue and improves downstream decision quality.  
The result holds uniformly across all calibrated radii $R(\alpha)$.

Together, these results demonstrate that CREDO provides conservative risk estimates (validity), converges to the true risk under reasonable conditions (consistency), and benefits from generative sampling to avoid excessive conservativeness (high TPR). These properties ensure both theoretical rigor and practical utility in decision support applications.

\section{Computational Implementation}
\label{sec:computational}

The theoretical analysis in the previous section establishes that CREDO provides conservative risk estimates with strong finite-sample guarantees. However, realizing these guarantees in practice requires efficiently solving \eqref{eq:alpha-raw} for each decision $z$ and generated sample $\hat{y}$. This presents a fundamental computational challenge as we must verify whether a calibrated conformal ball $\mathcal{C}(x;\alpha)$ is entirely contained within the inverse feasible region $\pi^{-1}(z)$, which is NP-hard in general.

This section develops practical algorithms to address this challenge. We begin by reformulating the set containment constraint into a standard optimization problem (Section~\ref{sec:reformulation}). Subsequently, this allows us to derive efficient solution strategies for two important problem classes (Section~\ref{sec:linear}): ($i$) linear programs (LPs), where we derive closed-form expressions, and ($ii$) general convex problems, where we employ gradient-based approximations. These computational insights enable practitioners to deploy CREDO for real-world decision support while maintaining the theoretical guarantees established in Section \ref{sec:theory}.
The proofs in this section are deferred to \Cref{app:proof}.

\subsection{Reformulating the Set-Containment Constraint}\label{sec:reformulation}

We begin by noticing that the set containment condition $\mathcal{C}(x;\alpha) \subseteq \pi^{-1}(z)$ is equivalent to requiring that the calibrated conformal ball does \emph{not} intersect the complement of the inverse feasible region:
\[
\mathcal{C}(x;\alpha) \subseteq \pi^{-1}(z)
\quad\Longleftrightarrow\quad
\mathcal{C}(x;\alpha) \cap \bigl( \pi^{-1}(z) \bigr)^c = \emptyset.
\]
Since $\mathcal{C}(x;\alpha)$ is an $\ell_2$ ball centered at $\hat y$, the right hand side is further equivalent to
\[
\underbrace{
\text{for all } y\in \bigl( \pi^{-1}(z) \bigr)^c, ~
\|y-\hat y\|_2\ge R(\alpha)
}_{(i)}
\qquad\text{if}~
\underbrace{
\hat y\in\pi^{-1}(z)
}_{(ii)}.
\]
Here, condition ($i$) requires that every scenario $y$ violating the optimality of decision $z$ must lie at least with distance $R(\alpha)$ away from $\hat y$;
condition ($ii$) states that the former condition is enforced only when $\hat y$ lies within of $\pi^{-1}(z)$. Otherwise, then no radius can guarantee containment, and the $\alpha$ is trivially mapped to one by design.
This reformulation allows us to circumvent the need for validating the original set containment relation defined in \Cref{eq:alpha-raw}, and derive a more principled computational representation, summarized as follows.
\begin{proposition}[Computation]
\label{prop:comp-eff}
Let $\hat y$ be a generated prediction and $\tilde \alpha$ be its estimated risk. 
Then
\begin{subequations}
    \label{eq:alpha-piecewise}
    \begin{align}
        \label{eq:alpha-case1}
        \tilde \alpha(z) &= 1,
        && \text{if } z \notin \pi_\epsilon(\hat y), \\[6pt]
        \label{eq:bilevel}
        \tilde \alpha(z) &=
        \sup_{y \in \mathcal{Y}}
        ~ R^{-1}\bigl(\|y-\hat y\|_2\bigr)
        \quad \text{s.t.}\quad
        f(z;y) > \min_{z' \in \mathcal{Z}} f(z';y) + \epsilon,
        && \text{if } z \in \pi_\epsilon(\hat y),
    \end{align}
\end{subequations}
where we expand the notation $\pi$ to $\pi_\epsilon$ to avoid ambiguity, and $R^{-1}(\cdot)$ is the inverse radius function:
\[
R^{-1}(l)
=
\min_{\alpha\in[0,1]}
\bigl\{\alpha : R(\alpha) \le l \bigr\}.
\]
\end{proposition}
Furthermore, the inverse radius functions for the three calibrated radii in Equations~\eqref{eq:p-radius}--\eqref{eq:inf-radius} of \Cref{prop:comp-eff} admit closed-form expressions, which makes solving for \eqref{eq:bilevel} a more explicit task. 
\begin{lemma}
\label{lem:three-inverse-function}
For the radii defined in Equations~\eqref{eq:p-radius}--\eqref{eq:inf-radius}, the corresponding inverse functions are:
\begin{align}
R_p^{-1}(l)
&=
\left(
1 - \frac{1}{n+1}
\left\lfloor 
\sum_{i=1}^n \mathbbm{1}\{L_i \le l\}
\right\rfloor\right)^+,
\label{eq:p-radius-inv}
\\[4pt]
R_e^{-1}(l)
&=
\left(
\frac{\sum_{i=1}^n L_i + l}{(n+1)l}\right)^+,
\label{eq:e-radius-inv}
\\[4pt]
R_\infty^{-1}(l)
&= 0,
\label{eq:inf-radius-inv}
\end{align}
where $(\cdot)^+ = \max\{0, \cdot\}$ denotes the rectified linear unit (ReLU) operator.
\end{lemma}

Additionally, we note that these results also shed light on the namesake of the Monte-Carlo radius $R_\infty$, which is summarized in the following remark.

\begin{remark}
    \label{rmk:monte-carlo}
    Combining \Cref{eq:inf-radius-inv} with \Cref{prop:comp-eff}, one can verify that the risk estimate computed with the Monte Carlo radius degenerates to the form:
    $$
    \hat \alpha(z) = \frac{1}{K} \sum_{k = 1}^K \mathbbm{1}\{ z \notin \pi(\hat y^{(k)})\}.
    $$
    Namely, the risk estimator is essentially a Monte Carlo estimator of whether the generated samples from the generative model $g(x)$ fall inside $\pi(\hat y^{(k)})$.
\end{remark}

\subsection{Computational Strategies for Specific Problem Classes}
\label{sec:linear} 

Leveraging \Cref{prop:comp-eff}, the computation of each individual risk estimate $\tilde \alpha(z)$ can be executed as two steps: 
($i$) Given a generated sample $\hat y$, we solve the original decision-making problem \eqref{eq:epsilon-solution-set} once to check if $z \in \pi(\hat y)$.
If not, we immediately set $\tilde \alpha(z)$ conservatively to one. Otherwise, we compute $\tilde \alpha(z)$ via solving \eqref{eq:bilevel}, which admits different strategies depending on the form of the objective $f(z;y)$ and the geometry of $\mathcal{Z}$.
In what follows, we analyze by specific problem classes, where we first provide a closed-form estimator for linear problems, and then introduce a heuristic approach applicable to general convex decision models.

\subsubsection{Linear Problems}

We first examine the special case in which the decision problem is a linear program (LP) of the form
\begin{equation}
    \label{eq:linear}
    f(z;y)
    \coloneqq
    \langle y, z \rangle
    \quad
    \text{s.t.}\quad
    A z \le b,
\end{equation}
where $A \in \mathbb{R}^{m \times d}$ and $b \in \mathbb{R}^m$ define a nonempty polyhedral feasible region.
We use $\mathcal{V}$ to denote the set of extreme points of this polytope, which can be solved from $A$ and $b$ using procedures such as the double description method \citep{motzkin1953double, fukuda1995double}.  

In this setting, we can derive a closed-form solution for the optimization problem \eqref{eq:bilevel} by leveraging $\mathcal{V}$ and the property that the optimal solution to an LP always occurs at an extreme point \citep{bertsimas1997introduction}.  
Specifically, by the fact that optimal solutions for LPs are always incurred at extreme points \citep{krein1940extreme}, the constraint in \eqref{eq:bilevel} can be equivalently expressed as
$$
f(z;y) > \min_{z' \in \mathcal{Z}} f(z';y) + \epsilon
\quad \Longleftrightarrow \quad 
\text{For all }v \in \mathcal{V}\setminus\{z\},
\quad 
\langle y, v - z \rangle - \epsilon < 0,
$$
Consequently, its feasible region is the union of a finite number of halfspaces, each associated with a direction vector $z-v$ and offset $-\epsilon$. 
Therefore, the nested optimization problem \eqref{eq:bilevel} reduces to evaluating a finite set of closed-form expressions, yielding an explicit formula for the risk estimate.

\begin{corollary}[Closed-Form Risk Estimation for Linear Programs]
\label{prop:closed-form}
Under \eqref{eq:linear}, the risk estimate $\tilde\alpha(z)$ for a single generated sample $\hat y$ admits the closed-form expression for \eqref{eq:bilevel}:
\[
\tilde \alpha(z) = \max_{v \in \mathcal{V} \setminus \{z\}} R^{-1}\left(
\frac{
\bigl|
\langle \hat y, z - v \rangle - \epsilon 
\bigr|
}{
\| z - v \|_2
}
\right),\quad~\text{if}~z \in \pi_\epsilon(\hat y),
\]
where we also expand the notation $\pi$ to $\pi_\epsilon$ to avoid ambiguity.
\end{corollary}

\subsubsection{Convex Problems}

In the linear setting, the success in deriving a closed-form estimate critically relies on the polyhedral description of the $\epsilon$-suboptimality region. However, for general convex decision problems, where we assume $\mathcal{Z}$ is convex and $f(z;y)$ is convex in each argument marginally, such a representation is unavailable.
Additionally, this also renders the constraint
$
f(z;y) - \min_{z' \in \mathcal{Z}}f(z';y) > \epsilon
$
to be potentially nonconvex in $(z,y)$, and cannot be directly handled by off-the-shelf convex solvers.  

We adopt a heuristic algorithm (\Cref{alg:convex}) that combines ideas from the difference-of-convex algorithm (DCA) \citep{tao1997convex} and coordinate descent \citep{wright2015coordinate} to solve for Problem~\eqref{eq:bilevel}.
By the monotonically decreasing property of $R^{-1}(\cdot)$, we begin by noticing that the original optimization problem can be reformulated as first solving the following single-level minimization problem
\begin{equation}
    \label{eq:single-level}
    \min_{y, z'}
    \|y - \hat y\|_2
    \quad \text{s.t.} \quad
    f(z;y) - f(z';y)
    > \epsilon
    ~~\text{and}~~
    z' \in \mathcal{Z}.
\end{equation}
and then applying the inverse radius function $R^{-1}(\cdot)$ to the optimal objective value.
To solve \eqref{eq:single-level}, which jointly minimizes over $(y, z')$, we adopt an alternating scheme that iteratively fixes one variable and solves the resulting marginal convex subproblem in the other.
The algorithm starts by initialize $y$ at the generated prediction $y^{(0)} = \hat y$.
For each iteration $t = 1,\dots,T$, the procedure:
($i$) Solve for $z^{(t)}$, which is computed as the optimal solution to the original decision-making problem with $y = y^{(t-1)}$. It represents the most feasible solution $z \in \mathcal{Z}$ at the current iteration, as one can verify that it maximizes the slack of the constraint in Problem~\eqref{eq:single-level}. 
($ii$) Solve for $y^{(t)}$, where we adopt a convex surrogate to the original constraint by approximating $f(z;y)$ with its first-order Taylor expansion, which can then be solved with standard solvers.
After $T$ iterations, the final iterate $y^{(T)}$ is used to form the approximate risk estimate
$$
\tilde \alpha(z)
\coloneqq
R^{-1}\left( \|y^{(T)} - \hat y \|_2\right), \quad~\text{if}~ z \in \pi(\hat y).
$$
In \Cref{sec:exp3}, we empirically validate this adopted approach by showing it is reliable for producing high-quality risk estimates, providing a computationally tractable solution to decision risk assessment tasks in broader convex settings.

\begin{algorithm}[!t]
\caption{
Alternating Optimization for Solving \eqref{eq:bilevel} in General Convex Settings
}
\label{alg:convex}
\begin{algorithmic}[1]
    \REQUIRE Prediction $\hat y$;
    Decision $z$;
    Tolerance level $\epsilon$.
    \STATE $y^{(0)} \leftarrow \hat y$;
    \FOR{$t \in \{1, \ldots, T\}$}
        \STATE $z^{(t)} \leftarrow \argmin_{z' \in \mathcal{Z}} f(z'; y^{(t - 1)})$;
        \STATE $\tilde f^{(t)} (z;y) \leftarrow f(z; y^{(t-1)}) + \left\langle \nabla_{y} f(z; y) \big|_{y = y^{(t-1)}} ~,~ y - y^{(t-1)} \right\rangle$;
        \STATE $y^{(t)} \leftarrow \argmin_{y \in \mathcal{Y}}
        \|y - \hat y\|_2
        \quad \text{s.t.} \quad
        \tilde f^{(t)} (z;y) - f(z^{(t)};y) > \epsilon$;
    \ENDFOR
    \RETURN $R^{-1}(\| y^{(T)} - \hat y \|_2)$.
\end{algorithmic}
\end{algorithm}

\section{Experiments}
\label{sec:experiments}

In this section, we evaluate the performance of CREDO through extensive numerical experiments.
Our results demonstrate that:
($i$) High-quality risk estimation (\Cref{sec:exp1}): CREDO produces risk estimates that can be tuned toward either conservativeness or accuracy through the choice of radius, with both aspects outperforming baseline models. This flexibility underscores its suitability for decision risk-assessment tasks.
($ii$) Risk-aware decision prescriptions (\Cref{sec:exp2}): The risk estimates produced by CREDO can be directly used to prescribe decisions that achieve consistent low risk, demonstrating CREDO’s reliability as a practical tool for risk-aware decision making.
($iii$) Effective modular design (\Cref{sec:exp3}): Each component of CREDO exhibits superior performance on its respective subtask compared with alternative ablation variants, highlighting the advantages of our modeling choices.
Additional details of the experiments are provided in \Cref{app:exp}.

\begin{figure}
    \FIGURE{
    \begin{subfigure}{\linewidth}
        \hspace{1ex}
        \includegraphics[width = 0.9\linewidth]{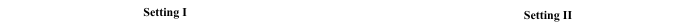}
        \includegraphics[width = 0.48\linewidth]{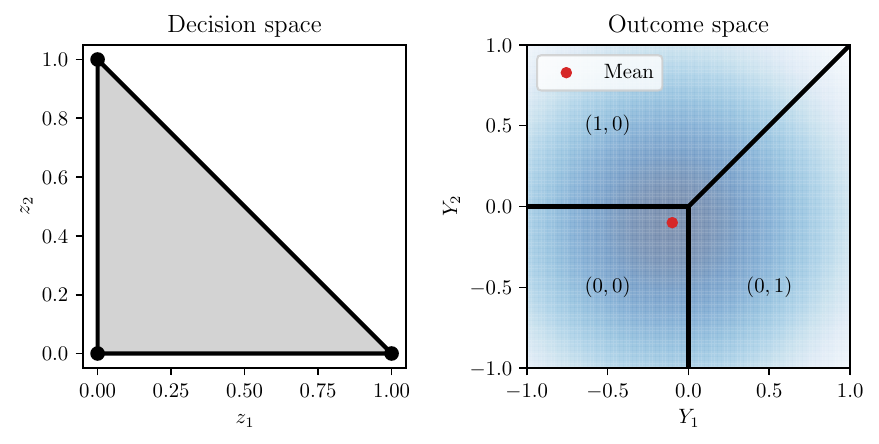}
        \includegraphics[width = 0.48\linewidth]{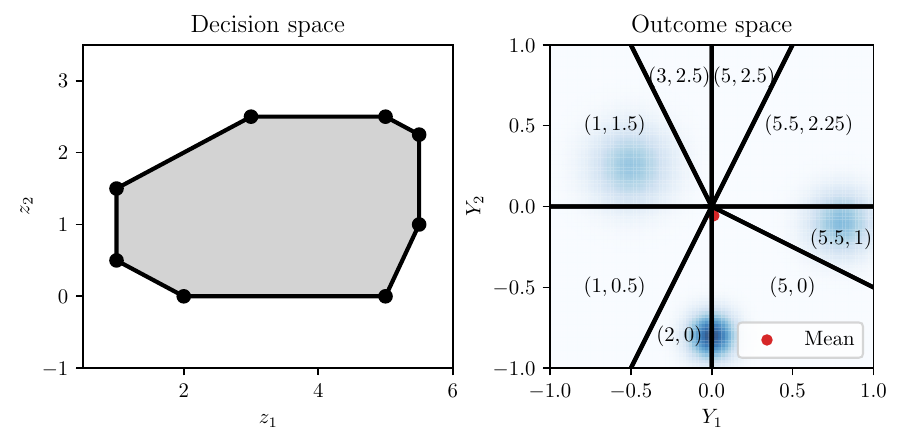}
    \end{subfigure}
    }
    {Illustration of settings I and II.}
    {The gray region represents the feasible region in the decision space, and the cones to the right are the corresponding inverse feasible regions in the outcome space. The blue shade denotes the density mass of~$Y$.
    \label{fig:synthetic-setting}
    }
\end{figure}

We test CREDO across a diverse set of optimization problems, including linear programming (LP), quadratic programming (QP), second-order conic programming (SOCP), integer programming (IP), and a penalized knapsack problem that uses semi-synthetic data from a real-world grid infrastructure investment problem (Real Data).
Specifically, we consider two stylized setups for the LPs, referred to as Setting I and Setting II, as illustrated in Figure~\ref{fig:synthetic-setting}. Setting I features a triangular feasible region with three vertices and a highly stochastic distribution of $Y$.
Setting II introduces a more complex octagonal feasible region with five vertices and a more dispersed, multimodal distribution of $Y$.
These two settings serve as running examples throughout our experiments due to their clarity and simplicity.
The variance of the synthetic data distribution is controlled by a hyperparameter $\sigma$.
Tabular results report the mean $\pm$ standard deviation across all trials. Boldface indicates the best-performing method, and underlining indicates the second-best within each setting.

\subsection{Risk Estimation Evaluation}
\label{sec:exp1}

In this subsection, we evaluate CREDO’s ability to produce risk estimates that flexibly trade off conservativeness and accuracy depending on the choice of radius, demonstrating that it outperforms baseline models along both dimensions.
To this end, we benchmark CREDO’s risk estimation performance against generic probability estimation baselines.
Note that the quantity of interest, $\mathbb{P}\{ z \in \pi(Y) \}$, can be interpreted as predicting, for a given test instance $(x, y)$, the probability that a prescribed decision remains $\epsilon$-optimal under $y$. This perspective naturally allows us to adapt standard regression and classification models as baselines for the decision risk-assessment task.

Specifically, we consider five representative baseline models:
($i$) SA: sample average of indicator functions computed from historic data;
($ii$) LR: logistic regression;
($iii$) NN: a neural network-based (multi-layer perceptron) classifier;
($iv$) QE: quantile estimator;
($v$) CP \citep{papadopoulos2002inductive, vovk2005algorithmic}: standard conformal prediction model using the upper prediction interval as its output.
These baseline models capture the types of prediction approaches that decision-makers are likely to adopt as main candidates for estimating decision risk.
We adopt two evaluation metrics for the decision risk estimate:
($i$) Validity: the ratio of trials where the estimated risk successfully upper bounds the true risk;
($ii$) MAE: the mean absolute error of the risk estimate compared to the true risk.
These two metrics quantify conservativeness and accuracy, which are the two crucial dimensions of risk estimate evaluation.
Additional details of the experimental settings are included in \Cref{app:exp}.

\Cref{tab:risk} summarizes the results.
For the first three baselines (SA, LR, NN), we observe low estimation error but poor validity. This is expected: these models are trained to maximize predictive accuracy, not conservativeness, and thus systematically underestimate risk.
In contrast, the last two baselines (QE and CP) exhibit higher validity but substantially larger errors. Although both approaches can enforce conservativeness by adjusting the quantile level $q$, the choice of $q$ is continuous in $[0,1]$ and lacks any principled mechanism to guarantee validity across problems or distributional shifts.
By comparison, CREDO achieves the best of both worlds. With $p$-value and $e$-value radii, CREDO attains $100\%$ validity across all settings; with the Monte Carlo radius, it yields highly accurate risk estimates with (near) minimal error.
Together, these results highlight the versatility of CREDO: it can be tuned to deliver either guaranteed conservativeness or high-precision estimation, offering a principled and robust framework for decision risk assessment.

\begin{table}[t]
\TABLE{
Evaluated metrics for different risk estimation methods across different optimization settings. 
}
{
\begin{adjustbox}{max width=1.0\linewidth}
\begin{tabular}{ccccccccccc}
\toprule
 & \multicolumn{2}{c}{LP Setting I} & \multicolumn{2}{c}{LP Setting II} & \multicolumn{2}{c}{QP ($\epsilon=0.1$)} & \multicolumn{2}{c}{SOCP ($\epsilon=0.2$)} & \multicolumn{2}{c}{IP ($\epsilon=0.3$)} \\
\cmidrule(lr){2-3} \cmidrule(lr){4-5} \cmidrule(lr){6-7} \cmidrule(lr){8-9} \cmidrule(lr){10-11} 
 & Validity ($\uparrow$) & MAE ($\downarrow$) & Validity ($\uparrow$) & MAE ($\downarrow$) & Validity ($\uparrow$) & MAE ($\downarrow$) & Validity ($\uparrow$) & MAE ($\downarrow$) & Validity ($\uparrow$) & MAE ($\downarrow$) \\
\midrule
SA & $0.53 \pm 0.50$ & $\bf 0.04 \pm 0.03$ & $0.56 \pm 0.44$ & $\bf 0.03 \pm 0.02$ & $0.43 \pm 0.48$ & $\bf 0.04 \pm 0.03$ & $0.47 \pm 0.48$ & $\bf 0.03 \pm 0.03$ & $0.41 \pm 0.46$ & $\bf 0.04 \pm 0.03$ \\
LR & $0.50 \pm 0.48$ & $0.06 \pm 0.04$ & $0.59 \pm 0.39$ & $0.03 \pm 0.02$ & $0.47 \pm 0.50$ & $0.05 \pm 0.04$ & $0.50 \pm 0.45$ & $0.06 \pm 0.05$ & $0.45 \pm 0.46$ & $0.07 \pm 0.04$ \\
NN & $0.50 \pm 0.45$ & $0.10 \pm 0.09$ & $0.39 \pm 0.38$ & $0.05 \pm 0.03$ & $0.63 \pm 0.46$ & $0.08 \pm 0.05$ & $0.40 \pm 0.48$ & $0.08 \pm 0.06$ & $0.48 \pm 0.49$ & $0.09 \pm 0.06$ \\
QE & $0.00 \pm 0.00$ & $0.62 \pm 0.13$ & $0.89 \pm 0.16$ & $0.15 \pm 0.06$ & $0.03 \pm 0.10$ & $0.60 \pm 0.10$ & $0.00 \pm 0.00$ & $0.56 \pm 0.00$ & $0.14 \pm 0.04$ & $0.53 \pm 0.06$ \\
CP & $\bf 1.00 \pm 0.00$ & $0.31 \pm 0.00$ & $\bf 1.00 \pm 0.00$ & $0.12 \pm 0.00$ & $\bf 1.00 \pm 0.00$ & $0.37 \pm 0.00$ & $\bf 1.00 \pm 0.00$ & $0.43 \pm 0.01$ & $\bf 1.00 \pm 0.00$ & $0.31 \pm 0.00$ \\
\textbf{CREDO} ($p$) & $\bf 1.00 \pm 0.00$ & $0.27 \pm 0.01$ & $\bf 1.00 \pm 0.00$ & $0.11 \pm 0.00$ & $\bf 1.00 \pm 0.00$ & $0.16 \pm 0.03$ & $\bf 1.00 \pm 0.00$ & $0.38 \pm 0.02$ & $\bf 1.00 \pm 0.00$ & $0.25 \pm 0.02$ \\
\textbf{CREDO} ($e$) & $\bf 1.00 \pm 0.00$ & $0.31 \pm 0.00$ & $\bf 1.00 \pm 0.00$ & $0.12 \pm 0.00$ & $\bf 1.00 \pm 0.00$ & $0.18 \pm 0.04$ & $\bf 1.00 \pm 0.00$ & $0.44 \pm 0.00$ & $\bf 1.00 \pm 0.00$ & $0.31 \pm 0.00$ \\
\textbf{CREDO} ($\infty$) & $0.50 \pm 0.50$ & \underline{$0.05 \pm 0.03$} & $0.53 \pm 0.43$ & \underline{$0.03 \pm 0.02$} & $0.60 \pm 0.48$ & \underline{$0.05 \pm 0.03$} & $0.53 \pm 0.49$ & \underline{$0.05 \pm 0.03$} & $0.47 \pm 0.47$ & \underline{$0.05 \pm 0.04$} \\
\bottomrule
\end{tabular}
\end{adjustbox}
}
{\label{tab:risk}}
\end{table}

\subsection{Decision Prescription Evaluation}
\label{sec:exp2}

In this part, we examine CREDO's performance when its outputted risk estimate is combined with the risk minimization criterion to prescribe decisions, where the candidate decision with the lowest risk is selected as the output.
we show that CREDO can be used to select decisions with consistently high confidence, demonstrating its effectiveness in guiding practical decision-making. 

We benchmark CREDO against four baselines: predict-then-optimize (PTO) \citep{bertsimas2020predictive}, robust optimization (RO) \citep{bertsimas2006robust}, smart predict-then-optimize (SPO+) \citep{elmachtoub2022smart}, and decision-focused learning (DFL) \citep{amos2017optnet}.
These methods are chosen for comparison as they represent widely used, risk-averse, data-driven approaches.
For CREDO's generative model design, we adopt four different variants: the Gaussian Mixture model (GMM) with one, three, and five components, as well as the variational autoencoder (VAE) \citep{kingma2013auto}.
We use \textit{empirical confidence ranking} as our primary evaluation metric: given a decision policy $\pi$ and a test dataset $\{(x_i, y_i)\}_{i=1}^m$, we apply $\pi$ to each input $x_i$ to generate predicted decisions $\{z_i\}_{i=1}^m$. We then compute the score $\sum_{i=1}^m \phi(z_i)$, where $\phi: \mathcal{Z} \to \mathbb{Z}_+$ is a mapping that maps each prediction to a discrete rank based on its frequency among the ground-truth optimal decisions $\{z_i^*\}_{i=1}^m$ in the test set. This metric is designed to capture a method’s tendency to select decisions that are most likely to be optimal---a direct representation of risk in decision-making contexts.

\begin{table}
    \TABLE
    {Evaluated empirical confidence ranking ($\downarrow$) for different methods across different optimization settings.
    }
    {
        \begin{adjustbox}{max width=1.0\linewidth}
        \begin{threeparttable}
        \begin{tabular}{ c c c c c c c c } 
        \toprule[1pt]
        \multirow{2}{*}{\textbf{Method}} & \multicolumn{3}{c}{Setting I} & \multicolumn{3}{c}{Setting II} & \multicolumn{1}{c}{Real Data} \\
        \cmidrule(lr){2-4}  
        \cmidrule(lr){5-7}  
        & $\sigma = 0.1$ & $\sigma = 1$ & $\sigma = 10$ & $\sigma = 0.1$ & $\sigma = 1$ & $\sigma = 10$ &  \\
        
        \midrule
        PTO    & $\bf 1.00 \pm 0.00$   & $2.76 \pm 0.59$     & $2.24 \pm 0.79$       & $3.55 \pm 0.50$       & \underline{$3.36 \pm 0.48$}       &   $2.04 \pm 1.65$ & \underline{$1.75 \pm  1.69$} \\
        RO     & $\bf 1.00 \pm 0.00$   & $2.98 \pm 0.14$     & $3.00 \pm 0.00$       & $4.99 \pm 0.10$       & $6.00 \pm 0.00$       &   $3.98 \pm 0.80$ & $3.00 \pm  1.29$ \\
        SPO+   & $\bf 1.00 \pm 0.00$   & $2.68 \pm 0.65$     & $2.02 \pm 0.82$       & $3.95 \pm 1.20$       & $4.67 \pm 1.56$       &   $3.56 \pm 1.50$ & $2.67 \pm  1.43$ \\
        DFL    & $2.44 \pm 0.64$       & $1.83 \pm 0.81$     & $2.06 \pm 0.79$       & $3.60 \pm 1.52$       & $3.96 \pm 2.07$       &   $3.66 \pm 2.48$ & $1.92 \pm  1.04$ \\
        \textbf{CREDO} (1-GMM)  & $1.94 \pm 0.87$       & $1.56 \pm 0.54$ & \underline{$1.49 \pm 0.50$}   & $3.74 \pm 0.98$   & $3.94 \pm 1.37$   & $2.02 \pm 1.41$ & $1.92 \pm  1.04$ \\
        \textbf{CREDO} (3-GMM)  & $1.75 \pm 0.77$       & $\bf 1.61 \pm 0.56$ & $\bf 1.48 \pm 0.52$   & $1.05 \pm 0.22$   & $\bf 1.00 \pm 0.00$   & $2.03 \pm 0.96$ & $\bf 1.75\pm  0.92$ \\
        \textbf{CREDO} (5-GMM)  & $1.89 \pm 0.87$       & $1.65 \pm 0.62$ & $1.54 \pm 0.52$   & \underline{$1.03 \pm 0.17$}   & $\bf 1.00 \pm 0.00$   & \underline{$1.92 \pm 0.89$} & $1.92 \pm  1.04$ \\
        \textbf{CREDO} (VAE)  & \underline{$1.01 \pm 0.10$}       & \underline{$1.61 \pm 0.58$} & $1.77 \pm 0.71$   & $\bf 1.00 \pm 0.00$   & $\bf 1.00 \pm 0.00$   & $\bf 1.06 \pm 0.24$ & $1.92 \pm  1.04$ \\
        
        \bottomrule[1pt]
        \end{tabular}
        \end{threeparttable}
        \end{adjustbox}
    }
    {\label{tab:decision-analysis}}
\end{table}

\Cref{tab:decision-analysis} presents the comparison results. It can be seen that CREDO achieves the smallest ranking metric value across most datasets, on average selecting the top two most likely decisions across all datasets. Though it might seem concerning that the in Setting I ($\sigma = 0.1$), PTO, RO, and SPO+ all achieve better performance than CREDO, this is because when $\sigma$ is small, the data becomes highly concentrated around the mean, rendering the problem nearly deterministic and can be best dealt with point-prediction baselines.
Across different generative model designs in CREDO, the 1-GMM performs significantly worse than the other generative model specifications. Among the remaining models, VAE and 5-GMM generally achieve slightly better performance than 3-GMM across most trials. This trend supports the intuition that more expressive generative models within CREOD lead to improved risk estimation, which further improves its capability to guide risk-aware decision making, especially under highly uncertain environments.

\subsection{Ablation Studies}
\label{sec:exp3}

\begin{figure}
    \FIGURE{
    \includegraphics[width=1.0\linewidth]{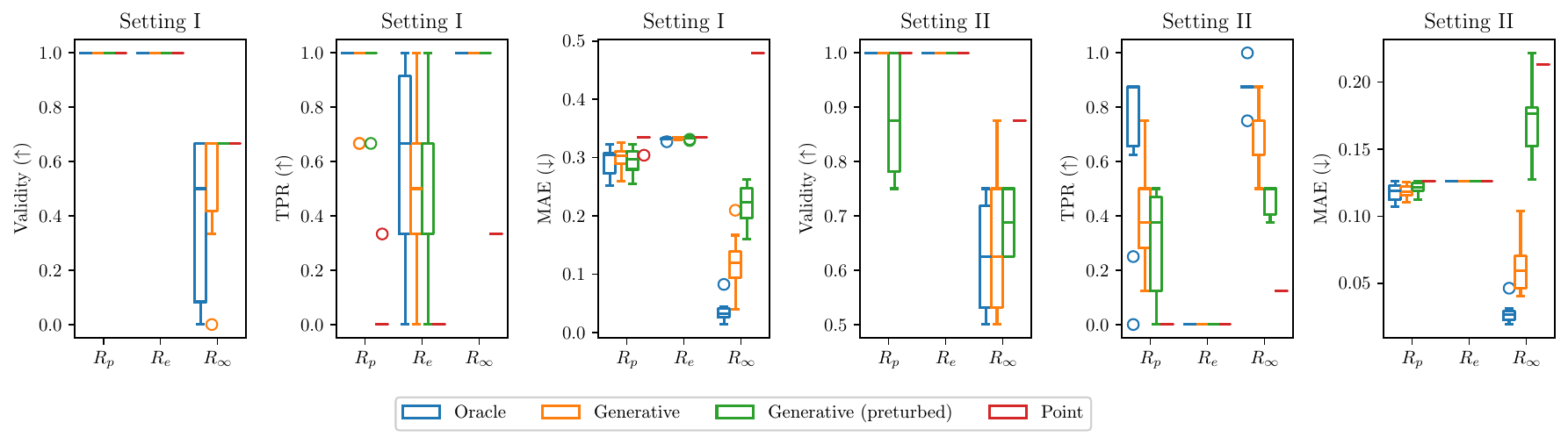}
    }
    {Comparison of CREDO performance for three different calibrated radii.\label{fig:exp1}}
    {}
\end{figure}

\begin{figure}
    \FIGURE{
    \begin{subfigure}{\linewidth}
        \centering
        \includegraphics[width=1.0\linewidth]{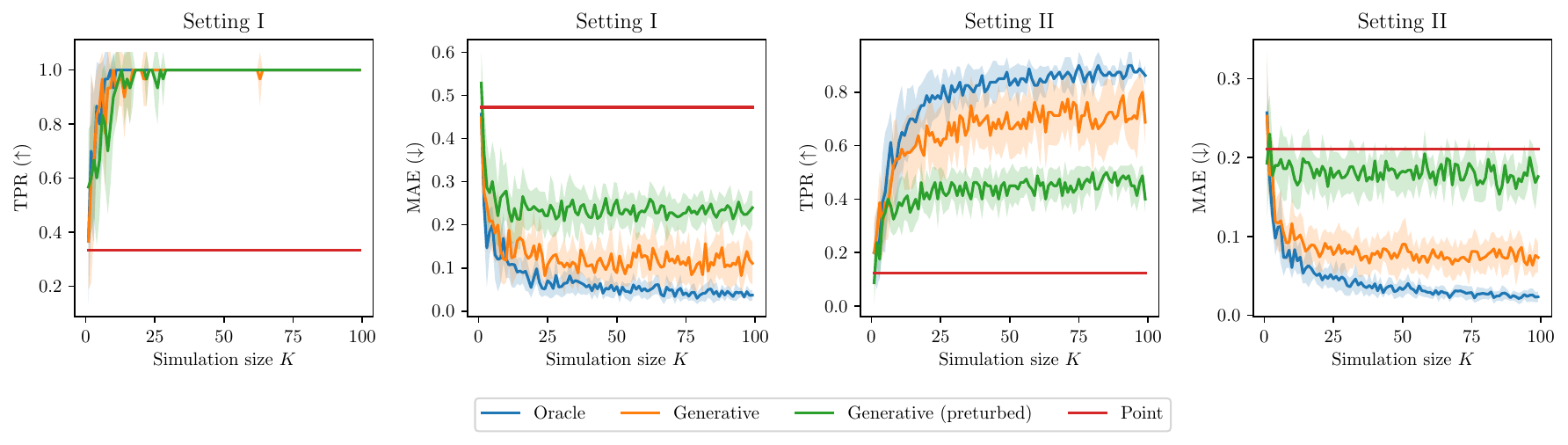}
        \\
        \includegraphics[width=1.0\linewidth]{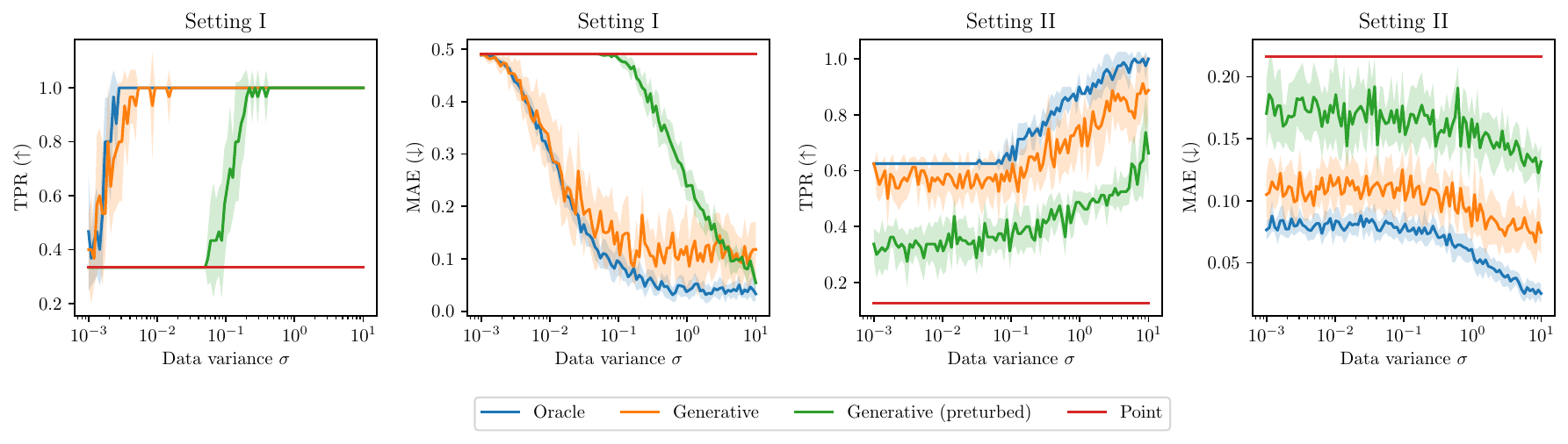}
    \end{subfigure}
    }
    {Comparison of CREDO performance for four different generative model configurations.\label{fig:exp2}}
    {\textit{Point} shows constant trend across different specifications of $K$ and $\sigma$, as they do not affect the fitting of the point model.
    }
\end{figure}

In this part, we examine ablation variants of CREDO through three controlled experiments that isolate the effects of radius design, generative modeling, and optimization procedures. These studies provide insight into CREDO’s sensitivity while also demonstrating the effectiveness of our design choices.

\Cref{fig:exp1} presents the evaluation results comparing different types of calibrated radii ($R_p$, $R_e$ and $R_\infty$) under different settings and generative model designs.
It can be seen that both $R_p$ and $R_e$ achieve a $100\%$ conservativeness rate across nearly all settings, while $R_\infty$ only achieves around $50\%$.
However, in terms of TPR and MAE, $R_p$ attains substantially better performance than both $R_e$ and $R_\infty$.
This pattern was also observed in \Cref{sec:exp1}, and here we further dig deeper to analyze the underlying insights:
($i$) Although for $R_p$, a formal theoretical proof of conservativeness is not available, it empirically demonstrates comparably strong conservativeness as $R_e$.
($ii$) Even though $R_\infty$ fails to ensure conservativeness, it achieves superior accuracy compared to both $R_p$ and $R_e$ by fully exploiting the generative model’s fidelity.
These insights highlight an essential trade-off when deploying CREDO: decision-makers must choose the calibrated radius according to their priorities.
If conservativeness is paramount, $R_p$ and $R_e$ should be preferred; if accuracy is more critical, $R_\infty$ is the better choice.
Additionally, a consistent pattern can be observed from \Cref{fig:exp1}: for both the TPR and accuracy metrics, $R_p$ achieves substantially higher values than $R_e$ across all settings. Since their conservativeness rates are nearly identical, this indicates that $R_p$ may serve as a more practical and effective choice than $R_e$, despite lacking a formal theoretical guarantee.
Consequently, $R_p$ can be interpreted as a balanced compromise between $R_e$ and $R_\infty$, suitable for scenarios where conservativeness and accuracy are of comparable importance.

For the second set of ablation studies, we implement CREDO with four different configurations of the underlying prediction (generative) model:
($i$) \textit{Oracle}: using the ground-truth data distribution as the generative model.
($ii$) \textit{Generative}: a three-component Gaussian Mixture model, fitted using the EM algorithm for $1 \times 10^2$ epochs.
($iii$) \textit{Generative (perturbed)}: same as \textit{Generative} except that the fitted mean of the components is perturbed with a small noise, which follows a $U(0.5, 1)$ distribution.
($iv$) \textit{Point}: point prediction model that captures the mean of the marginal distribution of $Y$.
The metrics used in this experiment follow those in the risk-estimation evaluation (\Cref{sec:exp1}), with additional inclusion of the true positive rate (TPR) in \eqref{eq:tpr}.

\Cref{fig:exp1} presents the evaluation results comparing different types of calibrated radii.
We observe that $R_p$ and $R_e$ achieve a $100\%$ conservativeness rate across nearly all settings, while $R_\infty$ maintains only about $50\%$.
However, in terms of TPR and accuracy, $R_p$ attains substantially higher values than both $R_e$ and $R_\infty$.
Two key insights emerge:
($i$) Although we lack a formal theoretical proof of conservativeness for $R_p$ as established for $R_e$ in \Cref{thm:val}, $R_p$ empirically demonstrates comparably strong empirical performance on conservativeness.
($ii$) Despite its failure to ensure conservativeness, $R_\infty$ achieves superior accuracy, as it behaves as an Monte Carlo estimator that fully exploits the generative model’s fidelity.
Overall, these findings highlight an essential trade-off when deploying CREDO: decision-makers must choose the calibrated radius according to their priorities.
If conservativeness is paramount, $R_e$ or $R_p$ should be preferred; if accuracy is more critical, $R_\infty$ is the better choice.
Additionally, a consistent pattern can be observed from \Cref{fig:exp1}: for both the TPR and accuracy metrics, $R_p$ achieves substantially higher values than $R_e$ across all settings. Since their conservativeness rates are nearly identical, this indicates that $R_p$ may serve as a more practical and effective choice than $R_e$, despite lacking a formal theoretical guarantee.
Consequently, $R_p$ can be interpreted as a balanced compromise between $R_e$ and $R_\infty$, suitable for scenarios where conservativeness and accuracy are of comparable importance.

\Cref{fig:exp2} presents the comparison results of CREDO risk estimation using different generative models.
Across all settings, it can be observed that as both $K$ and $\sigma$ increase, the three generative-based methods (\textit{Oracle}, \textit{Generative}, and \textit{Generative (perturbed)}) exhibit a clear improving trend, whereas the \textit{Point} model remains constant.
This pattern indicates that incorporating generative models within CREDO enhances the accuracy of risk estimation, particularly when the underlying data exhibit strong stochasticity (\ie, larger variance).
Moreover, the trends in \Cref{fig:exp2} generally follow the order \textit{Oracle} $>$ \textit{Generative} $>$ \textit{Generative (perturbed)} $>$ \textit{Point}.
Since both \textit{Generative (perturbed)} and \textit{Point} can be regarded as instances of misspecified models, this observation highlights that a well-trained generative model plays a crucial role in achieving accurate risk estimation.

\begin{table}
\TABLE
{Evaluated metrics for different optimization procedures solving \eqref{eq:bilevel} across different settings.}
{
\begin{adjustbox}{max width=1.0\linewidth}
\begin{tabular}{cccccccccccccccc}
\toprule
 & \multicolumn{3}{c}{LP Setting I} & \multicolumn{3}{c}{LP Setting II} & \multicolumn{3}{c}{QP} & \multicolumn{3}{c}{SOCP} & \multicolumn{3}{c}{IP} \\
\cmidrule(lr){2-4} \cmidrule(lr){5-7} \cmidrule(lr){8-10} \cmidrule(lr){11-13} \cmidrule(lr){14-16} 
 & Obj & Vio & Err & Obj & Vio & Err & Obj & Vio & Err & Obj & Vio & Err & Obj & Vio & Err \\
\midrule
GD & \underline{$0.44$} & $0.40$ & $0.45$ & \underline{$0.06$} & \underline{$0.07$} & \underline{$0.05$} & $0.60$ & $0.23$ & $0.50$ & $1.53$ & $0.13$ & $1.31$ & $0.22$ & $\bf 0.00$ & $0.19$ \\
BF & $0.66$ & $\bf 0.00$ & \underline{$0.43$} & $0.11$ & $\bf 0.00$ & $0.10$ & \underline{$0.52$} & $\bf 0.00$ & \underline{$0.34$} & \underline{$0.66$} & $\bf 0.00$ & \underline{$0.44$} & \underline{$0.11$} & $\bf 0.00$ & \underline{$0.09$} \\
RS & $0.89$ & $\bf 0.00$ & $0.66$ & $0.12$ & $\bf 0.00$ & $0.11$ & $0.70$ & $\bf 0.00$ & $0.52$ & $0.91$ & $\bf 0.00$ & $0.69$ & $0.12$ & $\bf 0.00$ & $0.10$ \\
RG & $5.21$ & \underline{$0.20$} & $4.98$ & $1.35$ & $0.17$ & $1.34$ & $4.81$ & $0.17$ & $4.63$ & $5.21$ & $0.17$ & $4.99$ & $1.35$ & $\bf 0.00$ & $1.32$ \\
\textbf{CREDO} & $\bf 0.23$ & $\bf 0.00$ & $\bf 0.00$ & $\bf 0.01$ & $\bf 0.00$ & $\bf 0.00$ & $\bf 0.10$ & \underline{$0.10$} & $\bf 0.08$ & $\bf 0.14$ & \underline{$0.03$} & $\bf 0.08$ & $\bf 0.00$ & $\bf 0.00$ & $\bf 0.02$ \\
\bottomrule
\end{tabular}
\end{adjustbox}
}
{\label{tab:opt}}
\end{table}

In the third set of experiments, we evaluate the performance of different optimization procedures solving Problem~\eqref{eq:bilevel}.
We use three general metrics:
($i$) Obj: the solution's objective value averaged across trials;
($ii$) Vio: the average ratio of solutions that violate the optimization constraint.
and ($iii$) Err: the average distance between the solution and the ground-truth solution.
The ground truth solution is computed from the closed-form solution under the linear setting, and approximated using a highly granular brute-force enumeration strategy in nonlinear settings.
We include five baselines:
($i$) GD: a generic gradient descent algorithm solving the optimization by penalizing the objective function with the constraint;
($ii$) BF: brute force enumeration, where its discretization resolution is set to be the same as the number of iterations $T$ for other methods;
($iii$) RS: random search algorithm, which is similar to BF except that the enumeration is done via sampling from a standard Gaussian distribution;
($iv$) RG: random guess, which is equivalent to a one-step RS.
These baselines are chosen as they are to the best of our knowledge, the few tractable procedures that can effectively handle Problem~\eqref{eq:bilevel}.
To ensure fair comparison, all the above optimization baselines are tuned with the number of iterations and/or epochs such that they take approximately the same time to run as our proposed method.

\Cref{tab:opt} reports the comparison of optimization procedures. Across all settings, CREDO consistently achieves low error. In the linear setting, CREDO attains zero error because the corresponding closed-form estimator recovers the exact solution.
For nonlinear convex settings, CREDO continues to exhibit low overall error, attributable to its ability to balance achieving a small objective value (Obj) while maintaining a low violation ratio (Vio).
Together, these results demonstrate that CREDO’s optimization procedure effectively handles both linear and nonlinear problems, supporting its efficacy and suitability for decision risk–assessment tasks and yielding the strong risk-estimation performance observed in \Cref{sec:exp1}.

\section{Discussion and Conclusions}
\label{sec:conclusion}

We introduced CREDO, a distribution-free framework for decision risk assessment that fundamentally shifts the paradigm from prescribing decisions to evaluating their reliability under uncertainty. By combining inverse optimization with conformal prediction, CREDO provides calibrated estimates of the probability that any candidate decision may be suboptimal, enabling practitioners to audit both human-proposed and algorithm-generated decisions with statistical guarantees.

Our theoretical analysis establishes that CREDO achieves conservative risk estimates, providing expectation-wise validity even under model misspecification. The framework's consistency properties ensure that these bounds tighten asymptotically as more data becomes available, while the true positive rate analysis demonstrates that generative sampling effectively reduces excessive conservativeness. These properties position CREDO as both theoretically rigorous and practically useful, addressing the long-standing challenge of quantifying decision stability without distributional assumptions.

The computational strategies we developed make these theoretical guarantees accessible in practice. For linear programs, our closed-form solution leverages the polytope structure to evaluate risk efficiently through vertex enumeration. For general convex problems, our gradient-based approach using differentiable optimization layers provides a practical approximation scheme. While computational complexity increases with problem scale and the number of generative samples, the modular design allows decision makers to balance computational cost against the desired level of risk assessment fidelity.

Our empirical evaluation reveals important trade-offs that emerge when deploying CREDO. The choice of calibrated radius fundamentally determines the balance between conservativeness and accuracy. The e-value approach offers rigorous post-hoc validity guarantees, but alternatives such as the p-value variant provide tighter risk estimates at the cost of weakened validity. Selecting an appropriate radius thus requires balancing the trade-off between validity and informativeness, ensuring that decision support remains both safe and actionable. This flexibility allows practitioners to select the variant that best matches their risk tolerance and application requirements. Moreover, our experiments demonstrate that CREDO's effectiveness scales with the quality of the underlying generative model. While validity is maintained even under misspecification, the informativeness of risk estimates depends critically on how well the generative model captures the true parameter distribution.

The framework's applicability extends to operational domains where parameter uncertainty significantly impacts decision quality. For example, for inventory management systems, CREDO can evaluate the robustness of reorder point policies and safety stock levels against demand variability and lead time uncertainty, providing quantitative assessments of when established inventory rules may fail to minimize total costs. Rather than prescribing singular optimal solutions, our framework provides decision-makers with risk profiles that characterize solution stability across the parameter space, enabling more informed trade-offs between expected performance and robustness. This paradigm shift from deterministic optimization to risk-aware assessment is particularly valuable in regulated industries where decision justification and risk documentation are required, as CREDO's calibrated guarantees provide auditable evidence of decision quality under uncertainty.

Several limitations of the current framework suggest directions for future research.
($i$) Overconservativeness with high-dimensional $\mathcal{Y}$: 
CREDO’s risk estimates become increasingly loose as the dimension of $\mathcal{Y}$ grows due to the curse of dimensionality---conformal sets expand in volume with dimension even at fixed coverage. While this preserves validity, it can reduce practical usefulness in decision-making tasks with many uncertain components. Future work may explore more flexible prediction-set geometries to improve efficiency \citep{izbicki2022cd, zheng2024optimizing}.
($ii$) Lack of convergence guarantees for the alternating algorithm:
Our alternating scheme in \Cref{alg:convex} works well empirically for solving \eqref{eq:single-level}, but a formal convergence analysis remains open. The problem structure suggests potential for adapting contraction-type or block coordinate descent results, offering a promising avenue for theoretical study.
($iii$) Human-in-the-loop evaluation: 
Because CREDO is intended to support human–algorithm collaboration, controlled user studies are useful to examine how real-world practitioners interact with its conservative risk assessments and whether these assessments improve decision quality relative to purely algorithmic recommendations.

To conclude, CREDO represents a step toward more transparent and accountable decision support systems. As machine learning increasingly influences critical decisions across all industry and public policy domains, the ability to quantify decision risk becomes essential for responsible deployment. The framework's distribution-free guarantees are particularly valuable in high-stakes settings where distributional assumptions are difficult to verify or where model misspecification could have severe consequences. By enabling algorithms to express uncertainty about their recommendations, CREDO facilitates a more nuanced collaboration between human expertise and machine intelligence, where each can contribute their respective strengths.


\ACKNOWLEDGMENT{Wenbin Zhou and Shixiang Zhu acknowledge partial support from the 2024 Block Center Seed Fund at Carnegie Mellon University and the 2024 GenAI Fellows program offered by the Tepper School of Business, Center for Intelligent Business.
Agni Orfanoudaki acknowledges support from the AI$^2$ Partnership Grant (from the UKRI and the AXA Insurance company), which facilitated this work.}

\bibliographystyle{informs2014}
\bibliography{ref}


\newpage
\begin{APPENDICES}


\section{Proofs}
\label{app:proof}

\subsection{Proof of \Cref{thm:val}}
\label{app:thm1}

For notation clarity, we will denote the $k$-th prediction set constructed by $\hat y^{(k)}$ as $\mathcal{C}^{k}(\cdot ; \cdot)$ in this and the following sections.
We begin by proving \textit{E-value post-hoc validity}, which is a key lemma to proving \Cref{thm:val}. 

\begin{lemma}[E-value post-hoc validity]
    \label{lem:val}
    Denote the prediction set constructed by $\hat y^{(k)}$ as $\mathcal{C}^{k}$.
    Then under \Cref{ass},
    $$
    \mathbb{P}\left\{ Y \in \mathcal{C}^{(k)}(X; \hat \alpha) \right\} \geq 1 - \mathbb{E} [ \hat \alpha ] , \quad \forall k = 1, \ldots, K.  
    $$
    where $\hat \alpha$ is some arbitrary function of $\{(X_i, Y_i)\}_{i = 1}^{n + 1}$. 
\end{lemma}

\begin{proof}{Proof of \Cref{lem:val}}
    When $\hat \alpha < 1 / (n + 1)$ or $\hat \alpha = 1$, by the definition of all three variants of $R(\alpha)$, the statement holds trivially.
    So we only need to consider the case when $1 > \hat \alpha \geq 1 / (n + 1)$.
    Recall that $\hat y_i$ denotes a single prediction generated from the calibration data $g(x_i)$.
    For any $k = 1, \ldots, K$, we begin by expanding the left-hand side:
    \begin{align*}
        \mathbb{P}\left\{ Y \notin \mathcal{C}^{(k)}(X; \hat \alpha) \right\}
        & = \mathbb{P}\left\{ \| \hat y^{(k)} - Y \|_2 > \frac{\sum_{i = 1}^n \| \hat y_i - Y_i \|_2 }{\hat \alpha (n + 1) - 1} \right\} \\
        & = \mathbb{P}\left\{ \hat \alpha (n + 1) \| Y - \hat y^{(k)} \|_2 > \sum_{i = 1}^n \| \hat y_i - Y_i \|_2 + \| \hat y^{(k)} - Y \|_2 \right\} \\
        & = \mathbb{P}\left\{ \hat \alpha > \frac{\sum_{i = 1}^n \| \hat y_i - Y_i \|_2 + \| \hat y^{(k)} - Y \|_2}{(n + 1) \| \hat y^{(k)} - Y \|_2} \right\} \\
        & = \mathbb{P}\left\{ \frac{(n + 1) \| \hat y^{(k)} - Y \|_2}{\sum_{i = 1}^n \| \hat y_i - Y_i \|_2 + \| \hat y^{(k)} - Y \|_2} > \frac{1}{\hat \alpha} \right\}.
    \end{align*}
    Denote the following random variables,
    \begin{align*}
        F_{i} & = \frac{(n + 1) \| \hat y_i - Y_i \|_2}{\sum_{i = 1}^n \| \hat y_i - Y_i \|_2 + \| \hat y^{(k)} - Y \|_2},
        \quad \forall i  = 1, \ldots, n, \\
        F_{n + 1} & = \frac{(n + 1) \| \hat y^{(k)} - Y \|_2}{\sum_{i = 1}^n \| \hat y_i - Y_i \|_2 + \| \hat y^{(k)} - Y \|_2}.
    \end{align*}
    One can prove that the following two conditions hold:
    \begin{align*}
        (a): \quad \quad  & \mathbb{E}[F_1 + \ldots + F_n + F_{n + 1}] = n + 1, \\
        (b): \quad \quad & \mathbb{E}[F_1] = \ldots  = \mathbb{E}[F_n] = \mathbb{E}[F_{n + 1}],
    \end{align*}
    where ($b$) holds by exchangeability (\Cref{ass}). Therefore, there is
    \begin{equation*}
        \mathbb{E}[F_{n + 1}] = 1.
    \end{equation*}
    Using this result, it can be derived that
    \begin{align*}
        \sup_{\tilde \alpha} \mathbb{E} \left[ \frac{\mathbb{P}(F_{n + 1} > 1 / \tilde \alpha)}{\tilde \alpha} \right]
        \leq \sup_{\tilde \alpha} \mathbb{E} \left[ \frac{\tilde \alpha \cdot \mathbb{E}[F_{n + 1}]}{\tilde \alpha} \right] 
        = \mathbb{E}[F_{n + 1}] = 1,
    \end{align*}
    where the inequality follows from Markov's inequality.
    Therefore, for any $\hat \alpha$ which may depend on the data $\{X_i, Y_i\}_{i = 1}^{n + 1}$, there is:
    $$
    \mathbb{E} \left[ \frac{\mathbb{P}(F_{n + 1} > 1 / \hat \alpha \mid \hat \alpha)}{\hat \alpha} \right]
    \leq \sup_{\tilde \alpha} \mathbb{E} \left[ \frac{\mathbb{P}(F_{n + 1} > 1 / \tilde \alpha)}{\tilde \alpha} \right]
    \leq 1
    $$
    Using a first-order Taylor expansion, the left-hand side of the first inequality is:
    \begin{equation}
        \label{eq:taylor}
        \mathbb{E} \left[ \frac{\mathbb{P}(F_{n + 1} > 1 / \hat \alpha \mid \hat \alpha)}{\hat \alpha} \right] \approx 
        \frac{\mathbb{E} \left[ \mathbb{P}(F_{n + 1} > 1 / \hat \alpha \mid \hat \alpha )\right]}{\mathbb{E} [\hat \alpha]}
    \end{equation}
    Assuming that this approximation is exact (see discussion in \Cref{rmk:taylor}), and by combining the two equations above, there is
    $$
    \mathbb{E} \left[ \mathbb{P}(F_{n + 1} > 1 / \hat \alpha \mid \hat \alpha) \right] \leq \mathbb{E}[\hat \alpha] \Longleftrightarrow
    \mathbb{P}\left\{ Y \in \mathcal{C}^{(k)}(X; \hat \alpha) \right\} \geq 1 - \mathbb{E} [ \hat \alpha ],
    $$
    where $\hat \alpha$ can probabilistically depend on $\{(X_i, Y_i)\}_{i = 1}^{n + 1}$ in arbitrary ways.
    \hfill \Halmos
\end{proof}

\begin{proof}{Proof of \Cref{thm:val}}
    We begin by noticing the following decomposition
    \begin{equation}
        \label{eq:cons-lower-bound}
        \mathbb{P}\left\{ z \in \pi(Y) \right\} = \mathbb{P}\left\{ Y \in \pi^{-1}(z ) \right\} \geq \frac{1}{K} \sum_{k = 1}^K \mathbb{P}\left\{ Y \in \mathcal{C}^{(k)}\left( X ; \tilde \alpha^{(k)}(z) \right) \right\}.
    \end{equation}
    The first equality follows from our problem reformulation (\Cref{lemma:risk-reformulation}). The second inequality holds due to the definition of $\tilde \alpha^{(k)}(z)$, which guarantees that the $k$-th generated CP region is always contained in $\pi^{-1}(z)$. By \Cref{lem:val}, there is: 
    $$
    \mathbb{P}\left\{ Y \in \mathcal{C}^{(k)}\left( X ; \tilde \alpha^{(k)}(z) \right) \right\} \geq 1 - \mathbb{E} \left[\tilde \alpha^{(k)}(z) \right].
    $$
    Therefore, combining this with \eqref{eq:cons-lower-bound} we obtain:
    $$
    \mathbb{P}\left\{ z \in \pi(Y) \right\} \geq 1 - \mathbb{E} \left[ \frac{1}{K} \sum_{k = 1}^K\tilde \alpha^{(k)}(z) \right] = 1 - \mathbb{E} \left[ \hat \alpha(z) \right].
    $$
    We conclude the proof for \Cref{thm:val}.
    \hfill \Halmos
\end{proof}

\begin{remark}[Discussion on the Taylor approximation]
\label{rmk:taylor}

We comment on the rationality of the first-order Taylor approximation used in \eqref{eq:taylor}.
To begin with, the approximation trick has been adopted in prior works \citep{gauthier2025values}, and argued that its error is small when the estimator $\hat \alpha$ is well concentrated around its mean.
Such a condition is usually satisfied in our setting. For example, when CREDO is deployed in a human-algorithm collaboration setting, the candidate decisions provided from the decision maker would be expected to be near optimal and should already enjoy a relatively small ground truth risk. This makes $\hat \alpha$ have a relatively small variance and well concentrated around its mean, which allows the Taylor approximation to be relatively tight.

Even when this condition does not hold, we can resort to an alternative way to account for the approximation error during risk estimator construction---deriving the approximation error and then manually offsetting the error in our risk estimator to achieve an exact conservativeness guarantee.
The derivation goes as follows: let $h(\hat \alpha) \coloneqq \mathbb{E}[\mathbb{P}(F_{n + 1} > 1 / \hat \alpha \mid \hat \alpha)]$, there is:
$$
\left| \mathbb{E}\left[ \frac{h(\hat \alpha)}{\hat \alpha} \right] - \frac{\mathbb{E}[h(\hat \alpha)]}{\mathbb{E}[\hat \alpha]} \right|
\leq \mathbb{E} [h(\hat \alpha)] \left( \mathbb{E} \left[ \frac{1}{\hat \alpha} \right] - \frac{1}{\mathbb{E}[\hat \alpha]} \right) + \sqrt{\operatorname{Var}(h(\hat \alpha)) \cdot \operatorname{Var}\left(\frac{1}{\hat \alpha}\right)}.
$$
The first term is the Jensen gap, and the second term is by the Cauchy-Schwarz inequality.
Assuming $\hat \alpha \in [\delta, 1]$ almost surely, then
$$
\mathbb{E} \left[ \frac{1}{\hat \alpha} \right] - \frac{1}{\mathbb{E}[\hat \alpha]} \leq \frac{1}{\delta^3} \operatorname{Var}(\hat \alpha), \quad \text{and} \quad
\operatorname{Var}\left(\frac{1}{\hat \alpha}\right) \leq \frac{1}{\delta^3} \operatorname{Var}\left(\hat \alpha\right).
$$
Plugging them into the first equation, we get
$$
\left| \mathbb{E}\left[ \frac{h(\hat \alpha)}{\hat \alpha} \right] - \frac{\mathbb{E}[h(\hat \alpha)]}{\mathbb{E}[\hat \alpha]} \right| \leq \frac{1}{\delta^3} \operatorname{Var}\left(\hat \alpha\right) + \frac{1}{2 \delta^2} \sqrt{\operatorname{Var}(\hat \alpha)}.
$$
Since for random variables bounded within $[\delta, 1]$, there is the following trivial upper bound:
$$
\operatorname{Var}(\hat \alpha) \leq \frac{1}{K^2} \sum_{k, k'} \operatorname{Cov}\left(I_k, I_k'\right) \leq \frac{1}{4},
$$
therefore, we can conclude that:
$$
\left| \mathbb{E} \left[ \frac{\mathbb{P}(F_{n + 1} > 1 / \hat \alpha \mid \hat \alpha)}{\hat \alpha} \right] -
\frac{\mathbb{E} \left[ \mathbb{P}(F_{n + 1} > 1 / \hat \alpha \mid \hat \alpha )\right]}{\mathbb{E} [\hat \alpha]} \right| \leq \frac{1}{4 \delta^3} + \frac{1}{4 \delta^2}.
$$
Plugging this result back into the proof of \Cref{thm:val}, we get
$$
\mathbb{P}\left\{ z \in \pi(Y) \right\} \geq 1 - \mathbb{E}[\hat \alpha(z)] - 1/4 (\delta^{-3} + \delta^{-2}).
$$
Therefore, one can take the final estimator as
\begin{equation}
    \label{eq:offset-estimator}
    \min\{ \hat \alpha(z) + 1/4 (\delta^{-3} + \delta^{-2}), 1\}
\end{equation}
so that exact conservativeness is achieved.
A trivial value that the user can take for $\delta$ is $1 / (n + 1)$, which is guaranteed by the design of the CREDO algorithm.
One can also manually tune the value of $\delta$ by modifying the calibrated radius as
\begin{equation*}
    R'(\alpha) = 
    \begin{cases}
        + \infty, & \text{if~} \alpha \in [ 0, \delta ), \\
        R(\alpha) & \text{if~} \alpha \in [\delta, 1 ), \\
        0  & \text{if~} \alpha = 1,
    \end{cases}
\end{equation*}
to achieve a tighter bound (\ie, smaller offset).
One can prove that as long as $\delta$ is chosen such that $\delta > 1/(n + 1)$, all theorems presented in the main text remain valid.
So one can safely adopt $R'(\alpha)$ as the conformalized radius and take \eqref{eq:offset-estimator} as the final estimator.
In the meantime, this procedure can be equivalently viewed as truncating the lower part of $\hat \alpha(z)$ at $\delta$, \ie, setting $\max\{\hat \alpha(z), \delta\}$ as the risk estimator, and then taking \eqref{eq:offset-estimator} as the final estimator.
\end{remark}

\subsection{Proof of \Cref{cor:p}}

\begin{proof}{Proof}
    Following the proof of Theorem 1 in \cite{barber2023conformal}, with the exception that all steps are conditioned on the $\sigma$-field generated by $\tilde \alpha^{(k)}(z)$, there is:
    $$
    \mathbb{P}\left[ Y \in \mathcal{C}^{(k)}(X ; \tilde \alpha^{(k)}(z)) \Big| \tilde \alpha^{(k)}(z) \right] \geq 1 - \alpha - \frac{\sum_{i = 1}^n  d_{\rm TV}^{(i)}(\hat \alpha(z)) }{n+1}.
    $$
    Here we implicitly utilized the fact that the $\sigma$-field generated by $\tilde \alpha^{(k)}(z)$ is the same across all $k$, as they are just different simulation trials based on the same input and generative model, therefore they are equal to the $\sigma$-field of $\hat \alpha(z)$.
    Plugging this equation back in \eqref{eq:cons-lower-bound}, we get
    $$
    \mathbb{P}\left\{ z \in \pi(Y) \right\} \geq 1 - \frac{1}{K} \sum_{k = 1}^K \mathbb{E} \left[ \tilde \alpha^{(k)}(z) + \frac{\sum_{i = 1}^n \mathbb{E} \left[ d_{\rm TV}^{(i)}(\hat \alpha(z)) \right]}{n+1} \right] = 1 - \mathbb{E} \left[ \tilde \alpha(z) \right] - \frac{\sum_{i = 1}^n  d_{\rm TV}^{(i)}(\hat \alpha(z)) }{n+1}.
    $$
    The expectations are taken with respect to all sources of randomness of $\tilde \alpha^{(k)}(z)$.
    \hfill \Halmos
\end{proof}

\subsection{Proof of \Cref{prop:consistency}}

\begin{proof}{Proof}
    By \Cref{rmk:monte-carlo}, when using $R_\infty$, the risk estimator is a Monte Carlo estimator: 
    $$
    \hat \alpha(z) = 1 - \frac{1}{K} \sum_{k = 1}^K \mathbbm{1} \left\{ \hat y^{(k)} \in \pi^{-1}(z ) \right\}.
    $$
    Therefore, for any $x \in \mathcal{X}$, conditioning on $X = x$ and given $f(x) = \hat{\mathcal{P}}_{Y|X=x}$, there is
    \begin{align*}
        |\hat \alpha(z) - \mathbb{P}\{z \in \pi(Y) \mid X = x\}|
        & = 1 - \frac{1}{K}\sum_{k = 1}^K \left( \mathbbm{1} \left\{ \hat y^{(k)} \in \pi^{-1}(z) \right\} - \mathbb{P}\{z \in \pi(Y) \mid X = x\} \right) - 1 \\
        & = \frac{1}{K}\sum_{k = 1}^K \left( \mathbbm{1} \left\{ \hat y^{(k)} \in \pi^{-1}(z ) \right\} - \mathbb{P}\{z \in \pi(Y) \mid Y \sim \mathcal{P}_{ Y | X = x}\} \right) \\
        & = \frac{1}{K}\sum_{k = 1}^K \Big( \underbrace{\mathbbm{1} \left\{ \hat y^{(k)} \in \pi^{-1}(z ) \right\} - \mathbb{P}\{z \in \pi(Y) \mid Y \sim \hat{\mathcal{P}}_{Y | X = x}\}}_{A_k} \Big)  +  \\
        & \quad  \frac{1}{K} \sum_{k = 1}^K \Big( \underbrace{\mathbb{P}\{z \in \pi(Y) \mid Y \sim \hat{\mathcal{P}}_{Y | X = x}\} - \mathbb{P}\{z \in \pi(Y) \mid Y \sim \mathcal{P}_{Y | X = x}\}}_{B_k} \Big) \\
        & = \frac{1}{K} \sum_{k = 1}^K A_k + \frac{1}{K} \sum_{k = 1}^K B_k.
    \end{align*}
    For the first term, since: ($i$) $A_k$ are i.i.d. random variables, as $\hat y^{(k)}$ are independently generated from the same generative model given $x$; ($ii$) $\mathbb{E}[A_k \mid X = x] = 0$, ($iii$) $A_k$ has finite variance, therefore by the central limit theorem, there is $\frac{1}{K} \sum_{k = 1}^K A_k = O_p(K^{-1/2})$. 
    For the second term, note that by the assumption
    $$
    T_k^{(2)} = \mathbb{P}\{Y \in \pi^{-1}(z) \mid Y \sim \hat{\mathcal{P}}_{Y | X = x}\} - \mathbb{P}\{Y \in \pi^{-1}(z) \mid Y \sim \mathcal{P}_{Y | X = x}\} \leq d_{\rm TV}(\hat{\mathcal{P}}_{Y| X = x}, \mathcal{P}_{Y|X = x}) \leq \delta,
    $$
    where the first inequality follows from the definition of total variation distance. Therefore, we conclude that
    $$
    |\hat \alpha(z) - \mathbb{P}\{z \in \pi(Y) \mid X = x\}| = O_p(K^{-1/2}) + \delta.
    $$
    This completes the proof.
    \hfill \Halmos
\end{proof}

\subsection{Proof of \Cref{prop:tpr}}

\begin{proof}{Proof}
    For notation simplicity, denote the sets in the numerator and the denominator of TPR defined in \eqref{eq:tpr} as:
    \begin{align*}
        A & = \{ z \in \mathcal{Z} \mid \alpha(z)< 1 \text{ and } \hat \alpha(z) < 1 \}, \\
        B & = \{ z \in \mathcal{Z} \mid \alpha(z) < 1 \}.
    \end{align*}
    For simplicity, we assume that $B$ (therefore $A$) is a finite set, \ie, there is only a finite set of decisions that have ground-truth risk smaller than one (though note that \Cref{prop:tpr} and its proof naturally extend to the infinite case by replacing the counting measure ``\#'' with continuous measures, such as Lebesgue measure defined within the decision space $\mathcal{Z}$).
    Since $B$ is irrelevant to $K$, we begin by expanding the following expression:
    \begin{align*}
        \mathbb{E}\left[\# A \right]
        & = \mathbb{E} \left[ \sum_{z \in \mathcal{Z}} \mathbbm{1}\left\{ \hat{\alpha}(z) < 1 
        ~\text{and}~
        \alpha(z) < 1 \right\}
        \right] \\
        & = \mathbb{E} \left[ \sum_{z \in B} \mathbbm{1}\left\{ \hat{\alpha}(z) < 1 \right\} \right] \\
        & = \mathbb{E} \left[ \sum_{z \in B} \left( 1 - \prod_{k = 1}^K 
        \left(
        \mathbbm{1} \left\{ \hat{y}^{(k)} \notin \pi^{-1}(z ) \right\}
        \right)
        \right) \right]
    \end{align*}
    The third equality results from the observation that the risk estimation is not one when at least one $\hat y^{(k)}$ falls within $\pi^{-1}(z)$.
    By the last expression, it can be seen that $\mathbb{E}\left[\# A \right]$ monotonically increases with $K$.
    Since $\text{TPR} = \mathbb{E}\left[\# A \right] / \mathbb{E}\left[\# B \right]$, it also monotonically increases with $K$.
    \hfill \Halmos
\end{proof}

\subsection{Proof of \Cref{prop:comp-eff}}

\begin{proof}{Proof}
    The first part of the proof resembles the main text.
    We begin by noticing that the set containment condition $\mathcal{C}(x;\alpha) \subseteq \pi^{-1}(z)$ is equivalent to requiring that the calibrated conformal ball does \emph{not} intersect the complement of the inverse feasible region:
    \[
    \mathcal{C}(x;\alpha) \subseteq \pi^{-1}(z)
    \quad\Longleftrightarrow\quad
    \mathcal{C}(x;\alpha) \cap \bigl( \pi^{-1}(z) \bigr)^c = \emptyset.
    \]
    Since $\mathcal{C}(x;\alpha)$ is an $\ell_2$ ball centered at $\hat y$, the right hand side is further equivalent to
    \begin{equation}
        \label{eq:boundary}
        \underbrace{
        \text{for all } y\in \bigl( \pi^{-1}(z) \bigr)^c, ~
        \|y-\hat y\|_2\ge R(\alpha)
        }_{(i)}
        \qquad\text{if}~
        \underbrace{
        \hat y\in\pi^{-1}(z)
        }_{(ii)}.
    \end{equation}
    Here, condition ($i$) requires that every scenario $y$ violating the optimality of decision $z$ must lie at least with distance $R(\alpha)$ away from $\hat y$;
    condition ($ii$) states that the former condition is enforced only when $\hat y$ lies within of $\pi^{-1}(z)$. Otherwise, then no radius can guarantee containment, and the $\alpha$ is trivially mapped to one by design.
    Therefore under ($ii$), \eqref{eq:alpha-raw} can be reformulated as
    \begin{equation}
        \label{eq:minmax}
        \min_{\alpha} \max_{y} \alpha
        \quad \text{s.t.} \quad 
        \|y-\hat y\|_2 \ge R(\alpha)
        \quad \text{and} \quad
        y \in \bigl( \pi^{-1}(z) \bigr)^c.
    \end{equation}
    Since by the monotonicity of $R$, there is
    $$
    \|y-\hat y\|_2 \ge R(\alpha) \quad\Longleftrightarrow\quad \alpha \geq R^{-1}\left(\|y-\hat y\|_2\right),
    $$
    so the first constraint in \eqref{eq:minmax} can be replaced, and the optimization can be rewritten as
    $$
    \min_{\alpha} \max_{y} \alpha
    \quad \text{s.t.} \quad 
    \alpha \geq R^{-1}\left(\|y-\hat y\|_2\right)
    \quad \text{and} \quad
    y \in \bigl( \pi^{-1}(z) \bigr)^c.
    $$
    In this problem. $\alpha$ can be viewed as a slack variable, and can be dropped so that the constraint becomes the objective:
    $$
    \max_{y} R^{-1}\left(\|y-\hat y\|_2\right)
    \quad \text{s.t.} \quad 
    y \in \bigl( \pi^{-1}(z) \bigr)^c.
    $$
    This turns the original minmax problem into a single-level maximization problem.
    Finally, by definition of $\pi^{-1}$, the constraint can be expanded as
    $$
    y \in \bigl( \pi^{-1}(z) \bigr)^c \quad\Longleftrightarrow\quad
    f(z ; y) > \min_{z' \in \mathcal{Z}} f(z'; y) + \epsilon.
    $$
    Plugging this into the previously derived optimization yields the desired statement. 
    \hfill \Halmos
\end{proof}

\subsection{Proof of \Cref{prop:closed-form}}

\begin{proof}{Proof}
    Using the linear assumption, we begin the derivation by expanding the constraint:
    \begin{equation*}
        f(z; y) > \min_{z' \in \mathcal{Z}} f(z'; y) + \epsilon 
        ~\Longleftrightarrow~  \min_{z' \in \mathcal{Z}}\langle y, z' - z \rangle + \epsilon < 0
    \end{equation*}
    Since $\mathcal{Z}$ is a compact set, by the Krein–Milman theorem \citep{krein1940extreme}, there is
    $$
    \min_{z' \in \mathcal{Z}}\langle y, z' - z \rangle = \min_{v \in \mathcal{V}}\langle y, v - z \rangle.
    $$
    Plugging this into \eqref{eq:bilevel} derived in \Cref{prop:comp-eff}, and using similar reformulation arguement as \eqref{eq:single-level}, we only need to solve the following optimization problem:
    $$
    \min_{y} \|y - \hat y\|_2
    \quad \text{s.t.} \quad
    \exists v \in \mathcal{V} \setminus \{z\}, \
    \langle y , v - z \rangle + \epsilon < 0
    $$
    This optimization finds the closest distance from a point $\hat y$ to \textit{any} halfspaces defined by norm vectors $z - v$ and offset $\epsilon$. By using the well-known point to halfspace distance formula, the optimal objective value of this optimization problem is:
    $$
    \min_{v \in \mathcal{V} \setminus \{z\}} \frac{ |\langle \hat y , z - v \rangle - \epsilon |}{\| z - v \|_2}.
    $$
    Therefore, the expression for the risk estimate is
    $$
    \tilde \alpha (z) = R^{-1} \left( \min_{v \in \mathcal{V} \setminus \{z\}} \frac{ |\langle \hat y , z - v \rangle - \epsilon |}{\| z - v \|_2} \right)
    =
    \max_{v \in \mathcal{V} \setminus \{z\}} R^{-1} \left( \frac{ |\langle \hat y , z - v \rangle - \epsilon |}{\| z - v \|_2} \right),
    \quad \text{if}~ z \in \pi_\epsilon(\hat y),
    $$
    where the second equality follows from the monotonic decreasing property of $R^{-1}$.
    This concludes the proof.
    \hfill \Halmos
\end{proof}

\begin{remark}[Indicator term]
    We note that under the linear case, the indicator term also admits the following closed-form:
    \begin{equation}
        \label{eq:ind-lin}
        \mathbbm{1} \left\{ \hat y^{(k)} \in \pi^{-1}(z) \right\} = \prod_{v \in \mathcal{V}} \mathbbm{1}\left\{ \langle \hat y^{(k)} , z - v \rangle \le  0 \right\}.
    \end{equation}
    To see why, we begin with the following expansion:
    \begin{equation*}
        \mathbbm{1}\left\{ y \in \pi^{-1}(z) \right\}
        = \mathbbm{1}\left\{ f(z; y) \leq \min_{z' \in \mathcal{Z}} f(z'; y) + \epsilon \right\} = \mathbbm{1} \left\{ \max_{z' \in \mathcal{Z}}\langle y, z - z' \rangle \le  \epsilon \right\}.
    \end{equation*}
    Using the Krein–Milman theorem from the proof above, we can derive that the right-hand side is equal to:
    $$
    \mathbbm{1} \left\{ \max_{z' \in \mathcal{Z}}\langle y, z - z' \rangle \le  \epsilon \right\} = \prod_{v \in \mathcal{V}} \mathbbm{1} \left\{ \langle y, z - v \rangle \le  \epsilon \right\}.
    $$
    Finally, setting $y = \hat y^{(k)}$ yields \eqref{eq:ind-lin}, which concludes the proof.
\end{remark}

\section{Additional Experiment Details}
\label{app:exp}

This section presents additional details of the numerical experiments that extend what has been described in the main text, and is organized as follows:
\Cref{app:spec} shows the resources and dependencies;
\Cref{app:opt} presents the detailed configurations for the optimization problems;
\Cref{app:baseline} describes the detailed configurations of the used baselines;
\Cref{app:metric} and \Cref{app:additional-result} describes the metrics and additional experiment results.

\subsection{Implementation Environments}
\label{app:spec}

All experiments were conducted on a machine running Windows 11, equipped with a 13th-generation Intel Core i7 CPU with 16 cores and 16 GB of RAM. No GPU acceleration was used in any of the experiments. The code is implemented in Python, and we list some key external packages:
\texttt{Scikit-learn} \citep{pedregosa2011scikit} is used for implementing Gaussian mixture models and some other statistical models that were used as baselines;
\texttt{CDD} \citep{fukuda1997cdd} is used for computing the vertices of polytopes to compute the closed-form solutions under the linear assumption;
\texttt{cvxpy} and \texttt{cvxpylayers} \citep{diamond2016cvxpy, cvxpylayers2019} are used to solve convex optimization problems and implement differentiable optimization layers;
\texttt{PyTorch} is used to build gradient-based baseline models.
Unless otherwise specified, we use the default parameter for all models in the package.
The complete code and dependencies are available in our codebase.

\subsection{Optimization problems}
\label{app:opt}

We describe the details of the six optimization problems that are featured in our experiments.
Throughout this section, we denote $\sigma$ as the component variance scale ($\sigma = 1$ by default), and $\mathbf{I}_d$ as the $d$-dimensional identity matrix.

\paragraph{LP (Setting I)} A linear programming problem with a triangular feasible region, defined as:
$$
\min_{z \in \mathbb{R}^2} Y_1 z_1 + Y_2 z_2
\quad \text{s.t.} \quad
z_1 + z_2 \leq 1 , z_1 \geq 0, z_2 \geq 0,
$$
where $(Y_1, Y_2)^\top$ is a Gaussian random vector with mean $(-1, -1)^\top$ and covaraince matrix $\sigma \cdot \mathbf{I}_2$. This optimization problem can be interpreted as a profit maximization task, where a manufacturer chooses the optimal production quantities $z_1$ and $z_2$ under a budget constraint, and $Y_1$ and $Y_2$ represent a risky market scenario that could still yield profit under favorable conditions where the expected revenue is negative.
Its feasible region can be more compactly represented in matrix form $\mathbf{A} z \leq \mathbf{b}$, where
$$
\mathbf{A} = 
\begin{pmatrix}
    1 & 1 \\
    -1    & 0 \\
    0 & -1
\end{pmatrix},
\quad 
\mathbf{b} = \begin{pmatrix}
1 \\
0 \\
0 
\end{pmatrix}
.
$$

\paragraph{LP (Setting II)} A linear programming problem that employs a more complex octagonal feasible region with five vertices and multimodal objective uncertainty,
defined as
$$
\min_{z \in \mathbb{R}^2} Y_1 z_1 + Y_2 z_2
\quad \text{s.t.} \quad
\mathbf{A} z \leq \mathbf{b},
$$
where the constraint matrix $\mathbf{A}$ and constraint vector $\mathbf{b}$ are defined as
$$
\mathbf A =
\begin{pmatrix}
-0.5 & 0 & -0.5 & 0.5 & 2 & 1 & 0 & -1 \\
-1   & -1 & 1 & 1 & -1 & 0 & 1 & 0
\end{pmatrix}^\top,
\quad
\mathbf b =
\begin{pmatrix}
-1 & 0 & 1 & 5 & 10 & 5.5 & 2.5 & -1
\end{pmatrix}^\top.
$$
The random vector $Y \in \mathbb{R}^2$ is drawn from a three-component Gaussian mixture distribution
\[
p(x) = \sum_{k=1}^{3} w_k ~ \mathcal{N}\left(x \mid \mu_k, ~\sigma \cdot \sigma_k^2 \mathbf{I}_2 \right),
\]
where the mixture weights are
$
\boldsymbol{w} = (0.3, 0.4, 0.3),
$
the component means $\boldsymbol \mu$ are defined as
\[
\mu_1 = \begin{pmatrix} 0.0, -0.8 \end{pmatrix}^\top, \quad
\mu_2 = \begin{pmatrix} -0.5, 0.25 \end{pmatrix}^\top, \quad
\mu_3 = \begin{pmatrix} 0.8, -0.1 \end{pmatrix}^\top,
\]
and the component variances $\boldsymbol \sigma$ are defined as
\[
\sigma_1^2 = (0.01)^2, \quad
\sigma_2^2 = (0.03)^2, \quad
\sigma_3^2 = (0.02)^2.
\]
Compared to the first setting, this setting allows us to assess risk assessment performance in more complex scenarios with multiple potentially optimal decisions.

\paragraph{QP} A quadratic programming problem defined as
$$
\min_{z \in \mathbb{R}^2} \frac{1}{2} z^\top Q z + \langle z, y \rangle
\quad \text{s.t.} \quad
\mathbf A z \leq \mathbf b,
$$
where the parameters are defined similar to \textit{LP (Setting I)}:
$$
Q = \begin{pmatrix}
    0.1 & 0 \\
    0 & 0.1 \\
\end{pmatrix},
\quad
\mathbf{A} = 
\begin{pmatrix}
    1 & 1 \\
    -1    & 0 \\
    0 & -1
\end{pmatrix},
\quad 
\mathbf{b} = \begin{pmatrix}
1 \\
0 \\
0 
\end{pmatrix}.
$$

\paragraph{SOCP}
A second-order conic programming problem, defined as
$$
\min_{z \in \mathbb{R}^2} ~\langle z, y \rangle
\quad \text{s.t.} \quad
0.1 \cdot \|z\|_2 \leq \mathbf A z - \mathbf b .
$$
where the parameters are defined similar to \textit{LP (Setting I)}:
$$
\mathbf{A} = 
\begin{pmatrix}
    1 & 1 \\
    -1    & 0 \\
    0 & -1
\end{pmatrix},
\quad 
\mathbf{b} = \begin{pmatrix}
1 \\
0 \\
0 
\end{pmatrix}.
$$

\paragraph{IP}
An integer programming problem, defined as
$$
\min_{z \in \mathbb{Z}^2} ~\langle z, y \rangle
\quad \text{s.t.} \quad
\mathbf A z - \mathbf b \leq 0.
$$
where the parameters are defined similar to \textit{LP (Setting II)}:
$$
\mathbf A =
\begin{pmatrix}
-0.5 & 0 & -0.5 & 0.5 & 2 & 1 & 0 & -1 \\
-1   & -1 & 1 & 1 & -1 & 0 & 1 & 0
\end{pmatrix}^\top,
\quad
\mathbf b =
\begin{pmatrix}
-1 & 0 & 1 & 5 & 10 & 5.5 & 2.5 & -1
\end{pmatrix}^\top.
$$

\paragraph{Real Data}

This setting considers a real-world power grid investment decision-making problem motivated by \citep{zhou2024hierarchical}.
A utility company based in Indianapolis, Indiana, has compiled detailed records of over 1,700 solar panel installations between 2010 and 2024, including the installation dates and affiliated grid components. With the renewable energy sector now at full scale, the management team anticipates a steady and significant monthly increase in solar adoption in the downtown area. In preparation for the incoming demand, they are planning targeted upgrades to grid-level inverters at $d = 4$ selected substations (we refer to them as Substation A to D) to minimize loss for grid failures subject to a limited budget.
This decision-making problem can be formulated as a penalized knapsack problem, defined as
\begin{equation*}
    \min_{ \mathbf{a} \in \{0, 1\}^d } \sum_{i = 1}^d \mathbbm{1}(Y_i \geq \tau_i) (1 - a_i) l_i \quad \text{s.t.} \quad  \sum_{i = 1}^d a_i c_i \leq b,
\end{equation*}
where $a_i \in {0,1}$ is a binary decision variable indicating whether substation $i$ is upgraded ($a_i=1$) or not ($a_i=0$); $Y_i$ is a random variable representing the monthly increase in solar panel installations connected to substation $i$; $c_i$ denotes the cost associated with upgrading substation $i$; $\tau_i$ is the capacity threshold, corresponding to the maximum allowable number of solar connections at substation $i$; $l_i$ represents the penalty incurred when the realized installations exceed $\tau_i$ without an upgrade; and $b$ denotes the total available budget.

The optimization problem above can be further relaxed into an LP problem:
$$
\max_{z \in \mathbb{R}^d} ~\langle Y, z \rangle
\quad \text{s.t.} \quad
\mathbf{A} z \leq \mathbf{b},
$$
where the parameters are defined as
\begin{equation*}
    Y = 
    \begin{pmatrix}
        {l_1}{\cdot (1 + e^{- \beta (Y_1 - \tau_1)})^{-1}} \\
        \ldots \\
        {l_d}{\cdot (1 + e^{- \beta (Y_d - \tau_d)})^{-1}} \\
    \end{pmatrix}, 
    \quad 
    z =
    \begin{pmatrix}
        1 - a_1 \\
        \ldots \\
        1 - a_d
    \end{pmatrix}
    ,
    \quad
    \mathbf{A} = 
    \begin{pmatrix}
        - \mathbf{c}^\top \\
        \mathbf{I}_d \\
        - \mathbf{I}_d
    \end{pmatrix},
    \quad 
    \mathbf{b} = 
    \begin{pmatrix}
        b - \mathbf{1}_d^\top \mathbf{c} \\
        \mathbf{1}_d \\
        \mathbf{0}_d
    \end{pmatrix}.
\end{equation*}
They are manually configured as follows:
The cost $c_i$ and loss $l_i$ for all substations equals one unit;
The capacity threshold $\tau_i$ is set to be the historical average solar panel monthly increment;
The budget $b$ is set to half the cost of upgrading all substations, allowing at most two out of four substations to be upgraded;
The smoothing parameter $\beta$ is set to $0.5$, which controls the sharpness of approximation to the thresholding function $\mathbbm{1}(Y_i \geq \tau_i)$.
This relaxation is used in our experiment so that the derived closed-form solution could be applied.

\begin{table*}[!t]
    \caption{Hyperparameter settings for CREDO.}
    \centering
    \begin{adjustbox}{max width=1.0\linewidth}
    \begin{threeparttable}
    \begin{tabular}{c l l} 
    \toprule[1pt]
    \textbf{Hyperparameter} & \textbf{Description} & \textbf{Default value} \\
    \midrule
    \textbf{$K$}    & Simulation size & $1 \times 10^2$ \\
     $\sigma$       & Component variance scale & $1$ \\
     $n$            & Calibration dataset size &    $1 \times 10^1$ \\
     $m$            & Training dataset size &       $1 \times 10^1$ \\
    \bottomrule[1pt]
    \end{tabular}
    \end{threeparttable}
    \end{adjustbox}
    \label{tab:param}
\end{table*}

\begin{table*}[!t]
    \caption{Summary table of selected metrics used in the experiments.}
    \centering
    \begin{adjustbox}{max width=0.9\linewidth}
    \begin{threeparttable}
    \begin{tabular}{p{0.2\textwidth} p{0.3\textwidth} p{0.5\textwidth}} 
    \toprule[1pt]
    \textbf{Metric Name} & \textbf{Description} & \textbf{Mathematical Formula} \\
    \midrule
    Validity
    &
    The percentage of decisions where the risk estimate satisfies the conservativeness guarantee in~\eqref{eq:risk-certificate}.
    &
    $$
    \frac{1}{|\mathcal{V}| } \sum_{z \in \mathcal{V}} \mathbbm{1} \left\{ \hat \alpha(z) \leq \alpha (z)\right\}
    $$
    \\
    MAE
    &
    The mean absolute error of the estimated compared to the true risks.
    &
    $$
    \frac{1}{|\mathcal{V}| } \sum_{z \in \mathcal{V}} \left|  \hat \alpha(z) - \alpha(z) \right|
    $$
    \\
    \bottomrule[1pt]
    \end{tabular}
    \end{threeparttable}
    \end{adjustbox}
    \label{tab:exp1-metrics}
\end{table*}

\subsection{Baselines}
\label{app:baseline}

In this part, we describe the baselines used in the experiments.
To control for variables, the baselines share the same set of hyperparameters as CREDO, summarized in \Cref{tab:param}.
Learning rates and their maximum epochs are set to $1 \times 10^{-2}$ and $1 \times 10^2$.
All optimization procedures are run for $10$ epochs.

\subsubsection{Baselines in \Cref{sec:exp1}}

\begin{itemize}
    \item[($i$)] \textit{SA (sample average)}: A naive sample average using observed historic data,
    $$
    \hat{\mathbb{P}}\{ z \not\in \pi(Y) \} = \frac{1}{n}\sum_{i = 1}^n \mathbbm{1}\{ z \not\in \pi(y_i)  \}
    $$
    Note that this estimator does not have any predictive power as it does not take $x$ as input, and does not guarantee conservativeness.
    \item[($ii$)] \textit{LR (logistic regression)}: Models the probability of $z$ falling within $\pi(Y)$ via the logistic regression model:
    $$
    \hat{\mathbb{P}}\{ z \not\in \pi(Y) \}
    =
    \frac{1}{1 + \exp\!\big(-(\theta^\top x + \theta_0)\big)},
    $$
    where $x$ denotes the feature vector and $(\theta,\theta_0)$ are learned parameters.
    \item[($iii$)] \textit{NN (neural network)}: Models the probability of $z$ falling within $\pi(Y)$ via a neural network classifier:
    $
    \hat{\mathbb{P}}\{ z \not\in \pi(Y) \}
    =
    \sigma\!\big(f_\theta(x)\big),
    $
    where $f_\theta(\cdot)$ denotes the neural network mapping parameterized by $\theta$, and $\sigma(t) = (1+\exp(-t))^{-1}$ is the logistic sigmoid function.
    \item[($iv$)] \textit{QE (quantile estimator)}: The $75\%$ empirical quantile of $\{\mathbbm{1}\{z \not\in \pi(y_i)\}\}_{i = 1}^n$. This is equivalent to quantile regression methods when $X$ is omitted.
    \item[($v$)] \textit{CP (conformal prediction)}: The naive conformal prediction method \citep{vovk2005algorithmic} applied to $\{ (x_i, \mathbbm{1}\{ z \in \pi(y_i)\}) \}_{i=1}^n$, taking the upper bound of the $25\%$ prediction interval as the final risk estimate:
    $$
    \hat{\mathbb{P}}\{ z \not\in \pi(Y) \}
    = 
    \hat{\mathbb{E}}[\mathbbm{1}\{ z \not\in \pi(Y)\} \mid X = x] + \hat Q_e\left(\frac{\lfloor (1 - 25\%) (n + 1)\rfloor}{n}\right),
    $$
    where $ \hat Q_e(\cdot)$ denotes the empirical quantile of the nonconformity scores (\ie, regression prediction residuals). The regression model is taken as SA (the sample average estimator).
\end{itemize}

\subsubsection{Baselines in \Cref{sec:exp2}}

\begin{itemize}
    \item[($i$)] \textit{{PTO} (predict-then-optimize)}: the standard two-stage predict-then-optimize approach \citep{bertsimas2020predictive}, which first predicts parameters and then solves the resulting optimization problem
    $
    \min_{z \in \mathcal{Z}} f(z; \hat Y),
    $
    where $\hat Y$ is a point estimate of $\mathbb{E}[ Y | X]$. Note that \texttt{PTO} is equivalent to stochastic optimization, since $X$ is omitted in our setting. $\hat Y$ is estimated using the sample average estimator from historic data.
    \item[($ii$)] \textit{{RO} (robust optimization)}: a minmax optimization problem
    $
    \min_{z \in \mathcal{Z}} \max_{y \in \mathcal{U}} f(z; y),
    $
    where the uncertainty set $\mathcal{U}$ is constructed using naive conformal prediction under the $\ell_\infty$ norm;
    \item[($iii$)] \textit{{SPO+} (smart predict-then-optimize)}: a predict-then-optimize method where the prediction model is trained using a surrogate loss that is convex and explicit \citep{elmachtoub2022smart};
    \item[($iv$)] \textit{{DFL} (decision-focused learning)}: prediction models are trained by directly optimizing the downstream objective through end-to-end differentiation of the optimization layer \citep{cvxpylayers2019, amos2017optnet}.
\end{itemize}
The baselines are selected based on the following criteria:
($i$) The ability to produce a decision that maximizes a (linear) objective function; and
($ii$) The capacity to handle randomness in the underlying optimization problem.

\subsubsection{Baselines in \Cref{sec:exp3}}

We describe the baselines used for comparing the optimization procedures:

\begin{algorithm}[!t]
\caption{
GD (Gradient Descent) for Solving \eqref{eq:bilevel} in Generla Convex Settings
}
\label{alg:grad}
\begin{algorithmic}[1]
    \REQUIRE Prediction $\hat y$;
    Decision $z$;
    Penalty parameter $\lambda$;
    Learning rate $\eta$;
    Tolerance level $\epsilon$.
    \STATE $y^{(0)} \leftarrow \hat y$ initialization;
    \FOR{$t \in \{1, \ldots, T\}$}
        \STATE $\psi_1^{(t - 1)} \leftarrow \nabla_y \min_{z' \in \mathcal{Z}} f(z'; y) \big|_{y = y^{(t -1)}}$ compute via differentiable optimization;
        \STATE $\psi_2^{(t - 1)} \leftarrow \nabla_y f(z; y) \big|_{y = y^{(t - 1)}}$;
        \STATE $\psi^{(t - 1)} \leftarrow \frac{y^{(t - 1)} - \hat y}{\| y^{(t - 1)} - \hat y \|_2} + \lambda \cdot \left( \psi_1 - \psi_2 + \epsilon \right)$ if constraint is positive else $\frac{y^{(t - 1)} - \hat y}{\| y^{(t - 1)} - \hat y \|_2}$;
        \STATE $y^{(t)} \leftarrow y^{(t - 1)} - \eta \cdot \psi^{(t - 1)}$;
    \ENDFOR
    \RETURN $\tilde \alpha(z) = R^{-1}(\| y^{(T)} - \hat y \|_2)$, if $z \in \pi_\epsilon(\hat y)$.
\end{algorithmic}
\end{algorithm}

\begin{itemize}
    \item[($i$)] \textit{GD (gradient descent)}:
    A gradient descent-based algorithm that approximates \eqref{eq:single-level} by minimizing its Lagrangian function, where its gradient is:
    $$
    \nabla_y \mathcal{L}(y, \lambda) = 
    \frac{ y - \hat y}{\| y - \hat y \|_2} + \lambda \cdot \nabla_y  \left[ \min_{z' \in \mathcal{Z}} f(z'; y) - f(z; y) + \epsilon \right]^+.
    $$
    Here, $\lambda$ is the regularization hyperparameter, which is set to $1 \times 10^3$.
    The term $\nabla_y \min_{z' \in \mathcal{Z}} f(z'; y)$ is computed through differentiable optimization layers \citep{cvxpylayers2019, amos2017optnet}.
    The pseudocode of GD is provided in \Cref{alg:grad}.
    \item[($ii$)] \textit{BF (brute force)}: A heuristic algorithm that first approximates the constraint space and decision space with gridded sets:
    $$
    \tilde{\mathcal{Y}} \approx  \{ y \in \mathcal{Y} \mid f(z;y) - f(z';y) > \epsilon \}
    \quad \text{and} \quad
    \tilde{\mathcal{Z}} \approx \mathcal{Z},
    $$
    then solves the finite optimization problem $\min_{y \in \tilde{\mathcal{Y}}} \min_{z' \in \tilde{\mathcal{Z}}} \|y - \hat y\|_2$. 
\end{itemize}
We note that \textit{RS (random search)} is similar to BF except that the gridded sets are constructed by drawing $T$ samples from a standard Gaussian distribution over the corresponding spaces.

\subsection{Metrics}
\label{app:metric}

\Cref{tab:exp1-metrics} summarizes three evaluation metrics used in our experiments.
The reporting standards for the metrics are their means $\pm$ standard deviations across $T = 100$ repeated trials.

\begin{algorithm}[!t]
\caption{Empirical Confidence Ranking}
\label{alg:rank}
\begin{algorithmic}[1]
    \REQUIRE True policy $\pi^*: \mathcal{Y} \to \mathcal{Z}$; Candidate policy $\pi: \mathcal{X} \to \mathcal{Z}$; Test dataset $\left\{ (x_i, y_i) \right\}_{i = 1}^\ell$.
    \STATE $\mathcal{Z}^* = \emptyset$
    \FOR{$i = \left\{ 1, \ldots, \ell \right\}$}
        \STATE $z_i^* \leftarrow \pi^*(y_i)$.
        \STATE $\mathcal{Z}^* \leftarrow \mathcal{Z}^* \cup \left\{ z_i^* \right\}$.
    \ENDFOR
    \FOR{$z \in \mathcal{V}(\theta)$}
        \STATE $h(z) \leftarrow$ the total occurence of decision $| \left\{ i \mid z^*_i = z \right\}|$.
    \ENDFOR
    \FOR{$i = \left\{ 1, \ldots, \ell\right\}$}
        \STATE $\hat z_i \leftarrow \pi(x_i)$.
        \STATE $\text{rank}_i \leftarrow$ ranking of estimated decision $|\left\{ z' \in \mathcal{V}(\theta) \mid h(z') \geq h(\hat z_i) \right\}|$
    \ENDFOR
    \RETURN Empirical confidence ranking $1/\ell \cdot \sum_{i = 1}^\ell \text{rank}_i$.
\end{algorithmic}
\end{algorithm}

\Cref{alg:rank} outlines the generic procedure for computing empirical confidence rankings. In the synthetic setting, this procedure is repeated over $T = 1 \times 10^2$ independent trials with $\ell = 1 \times 10^3$ test data points; In the real-world setting, the evaluation is performed rolling over $T = 12$ periods, spanning from 2010 to 2022. Each trial uses a two-year window (24 months) of data, which is sequentially split into training, calibration, and testing sets in an $8~:~8~:~8$ ratio.

\subsection{Additional Experiment Results}
\label{app:additional-result}

We augment the third set of experiments in \Cref{sec:exp3} with \Cref{tab:full-opt}, which is the full version of \Cref{tab:opt} with standard deviation included.
The interpretation of the table is the same as its main-text version.

\begin{table}
\centering
\TABLE
{Evaluated metrics for different optimization procedures solving \eqref{eq:bilevel}.}
{
\begin{subtable}{\linewidth}
\centering
\begin{adjustbox}{max width=\linewidth}
\begin{tabular}{cccccccccc}
\toprule
 & \multicolumn{3}{c}{LP Setting I} 
 & \multicolumn{3}{c}{LP Setting II} 
 & \multicolumn{3}{c}{QP} \\
\cmidrule(lr){2-4} \cmidrule(lr){5-7} \cmidrule(lr){8-10}
 & Obj & Vio & Err & Obj & Vio & Err & Obj & Vio & Err \\
\midrule
GD & \underline{$0.44 \pm 0.64$} & $0.40 \pm 0.20$ & $0.45 \pm 0.51$
   & \underline{$0.06 \pm 0.12$} & \underline{$0.07 \pm 0.06$} & \underline{$0.05 \pm 0.11$}
   & $0.60 \pm 0.43$ & $0.23 \pm 0.26$ & $0.50 \pm 0.33$ \\
BF & $0.66 \pm 0.42$ & $\bf 0.00 \pm 0.00$ & \underline{$0.43 \pm 0.37$}
   & $0.11 \pm 0.14$ & $\bf 0.00 \pm 0.00$ & $0.10 \pm 0.12$
   & \underline{$0.52 \pm 0.30$} & $\bf 0.00 \pm 0.00$ & \underline{$0.34 \pm 0.20$} \\
RS & $0.89 \pm 0.77$ & $\bf 0.00 \pm 0.00$ & $0.66 \pm 0.72$
   & $0.12 \pm 0.15$ & $\bf 0.00 \pm 0.00$ & $0.11 \pm 0.14$
   & $0.70 \pm 0.50$ & $\bf 0.00 \pm 0.00$ & $0.52 \pm 0.42$ \\
RG & $5.21 \pm 2.97$ & \underline{$0.20 \pm 0.22$} & $4.98 \pm 2.89$
   & $1.35 \pm 1.61$ & $0.17 \pm 0.15$ & $1.34 \pm 1.60$
   & $4.81 \pm 3.04$ & $0.17 \pm 0.17$ & $4.63 \pm 2.96$ \\
\textbf{CREDO}
   & $\bf 0.23 \pm 0.14$ & $\bf 0.00 \pm 0.00$ & $\bf 0.00 \pm 0.00$
   & $\bf 0.01 \pm 0.01$ & $\bf 0.00 \pm 0.00$ & $\bf 0.00 \pm 0.00$
   & $\bf 0.10 \pm 0.09$ & \underline{$0.10 \pm 0.15$} & $\bf 0.08 \pm 0.05$ \\
\bottomrule
\end{tabular}
\end{adjustbox}
\end{subtable}

\vspace{2ex}

\begin{subtable}{\linewidth}
\centering
\begin{adjustbox}{max width=\linewidth}
\begin{tabular}{ccccccc}
\toprule
 & \multicolumn{3}{c}{SOCP} & \multicolumn{3}{c}{IP} \\
\cmidrule(lr){2-4} \cmidrule(lr){5-7}
 & Obj & Vio & Err & Obj & Vio & Err \\
\midrule
GD & $1.53 \pm 0.92$ & $0.13 \pm 0.16$ & $1.31 \pm 0.81$
   & $0.22 \pm 0.27$ & $\bf 0.00 \pm 0.00$ & $0.19 \pm 0.24$ \\
BF & \underline{$0.66 \pm 0.42$} & $\bf 0.00 \pm 0.00$ & \underline{$0.44 \pm 0.29$}
   & \underline{$0.11 \pm 0.14$} & $\bf 0.00 \pm 0.00$ & \underline{$0.09 \pm 0.10$} \\
RS & $0.91 \pm 0.77$ & $\bf 0.00 \pm 0.00$ & $0.69 \pm 0.64$
   & $0.12 \pm 0.15$ & $\bf 0.00 \pm 0.00$ & $0.10 \pm 0.12$ \\
RG & $5.21 \pm 2.97$ & $0.17 \pm 0.17$ & $4.99 \pm 2.88$
   & $1.35 \pm 1.61$ & $\bf 0.00 \pm 0.00$ & $1.32 \pm 1.57$ \\
\textbf{CREDO}
   & $\bf 0.14 \pm 0.12$ & \underline{$0.03 \pm 0.10$} & $\bf 0.08 \pm 0.05$
   & $\bf 0.00 \pm 0.01$ & $\bf 0.00 \pm 0.00$ & $\bf 0.02 \pm 0.03$ \\
\bottomrule
\end{tabular}
\end{adjustbox}
\end{subtable}
}
{\label{tab:full-opt}}
\end{table}

\end{APPENDICES}

\end{document}